%% file: arxiv_submit_4996.tex
\documentclass[10pt,twocolumn,letterpaper]{article}

\usepackage{cvpr2020}
\usepackage{times}
\usepackage{epsfig}
\usepackage{graphicx}
\usepackage{amsmath}
\usepackage{amssymb}
\usepackage{booktabs}
\usepackage{adjustbox}
\usepackage{widetext}
\usepackage{tabu}
\usepackage{multirow}
\usepackage{caption}
\input{math_commands.tex}

\usepackage[pagebackref=true,breaklinks=true,letterpaper=true,colorlinks,bookmarks=false]{hyperref}

\cvprfinalcopy 

\def\cvprPaperID{4996} 

\providecommand{\tabularnewline}{\\}

\begin{document}

\title{Neural Pose Transfer by Spatially Adaptive Instance Normalization}

\author{\hspace{-0.7cm}Jiashun Wang$^{1*\ddagger}$\quad Chao Wen$^{1*\mathsection}$\quad Yanwei Fu$^{1 \dagger\ddagger}$\quad Haitao Lin$^{1}$\quad Tianyun Zou$^{1}$\quad Xiangyang Xue$^{1}$\quad Yinda Zhang$^{2}$\footnotemark[2]\\
\\
$^1$Fudan University \qquad $^2$Google LLC}

\twocolumn[{%
\renewcommand\twocolumn[1][]{#1}%
\maketitle
\vspace{-3.5em}
\begin{center}
    \centering
       \begin{tabular}{c}
    \hspace{-0.4in}
    \includegraphics[width=1.1\linewidth]{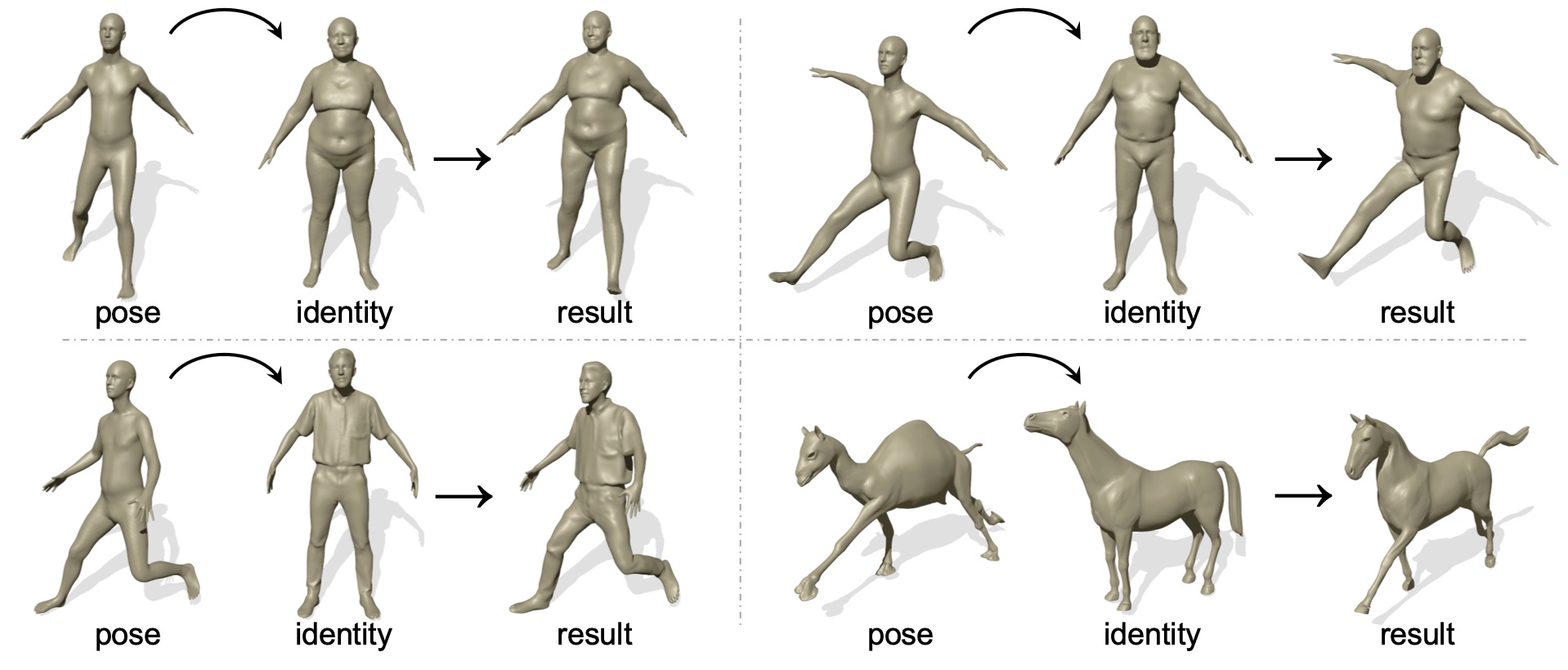}
        \end{tabular}
    \captionof{figure}{\textbf{Four groups of pose transfer examples.} Each visualization group consists of 3 meshes, input pose mesh, input identity mesh, and our result. For the first two groups, we show the pose mesh from SMPL~\cite{loper2015smpl}, the identity mesh from FAUST~\cite{Bogo_2014_CVPR} and our result, the identity mesh of the third group is from MG-dataset\cite{bhatnagar2019mgn}, the last group shows our pose transfer result from the animal dataset~\cite{sumner2004deformation}, please refer to supplementary materials for more details.}
    \label{fig:teaser}
\end{center}%
}]
{
  \renewcommand{\thefootnote}%
    {\fnsymbol{footnote}}
  \footnotetext[1]{indicates equal contributions.}
  \footnotetext[2]{indicates corresponding author.}
  \footnotetext[3]{Yanwei Fu and Jiashun Wang are with School of Data Science, and MOE Frontiers Center for Brain Science, Shanghai Key Lab of
Intelligent Information Processing Fudan University.}
 \footnotetext[4]{Chao Wen is with Academy of Engineering and Technology, and Institute of AI and Robotics, Fudan University}
}


\begin{abstract}
\vspace{-1em}
Pose transfer has been studied for decades, in which the pose of a source mesh is applied to a target mesh. 
Particularly in this paper, we are interested in transferring the pose of source human mesh to deform the target human mesh, while the source and target meshes may have different identity information.
Traditional studies assume that the paired source and target meshes are existed with the point-wise correspondences of user annotated landmarks/mesh points, which requires heavy labelling efforts. On the other hand, the generalization ability of deep models is limited, when the source and target meshes have different identities.
To break this limitation, we proposes the first neural pose transfer model that solves the pose transfer via the latest technique for image style transfer, leveraging the newly proposed component -- spatially adaptive instance normalization. 
Our model does not require any correspondences between the source and target meshes.
Extensive experiments show that the proposed model can effectively transfer deformation from source to target meshes, and has good generalization ability to deal with unseen identities or poses of meshes. Code is available at \href{https://github.com/jiashunwang/Neural-Pose-Transfer}{https://github.com/jiashunwang/Neural-Pose-Transfer}.
\end{abstract}

\vspace{-2em}
\section{Introduction}
Deformation transfer has been drawing consistent attention over decades and is enabling many applications.
For example, one can easily transfer the pose from the mesh of one person to another in the games and movies. However, It is very challenging when  there is a huge "shape gap" given very different identities of source and target meshes, as illustrated in Fig.~\ref{fig:teaser}.  
To make this feasible, previous works  demand re-enforcing the correspondence between source and target meshes,  additional information, such as  point-wise correspondence \cite{sumner2004deformation}, an auxiliary mesh \cite{xu2007gradient}, human key point annotations \cite{ben2009spatial}, skeleton pose \cite{chu2010example},  dense correspondence \cite{groueix20183d}, and so on.  Unfortunately, it is non-trivial,  and time-consuming to obtain such additional inputs  for deformation transfer. 

\begin{figure}[t]
\begin{center}

\includegraphics[width=1\linewidth]{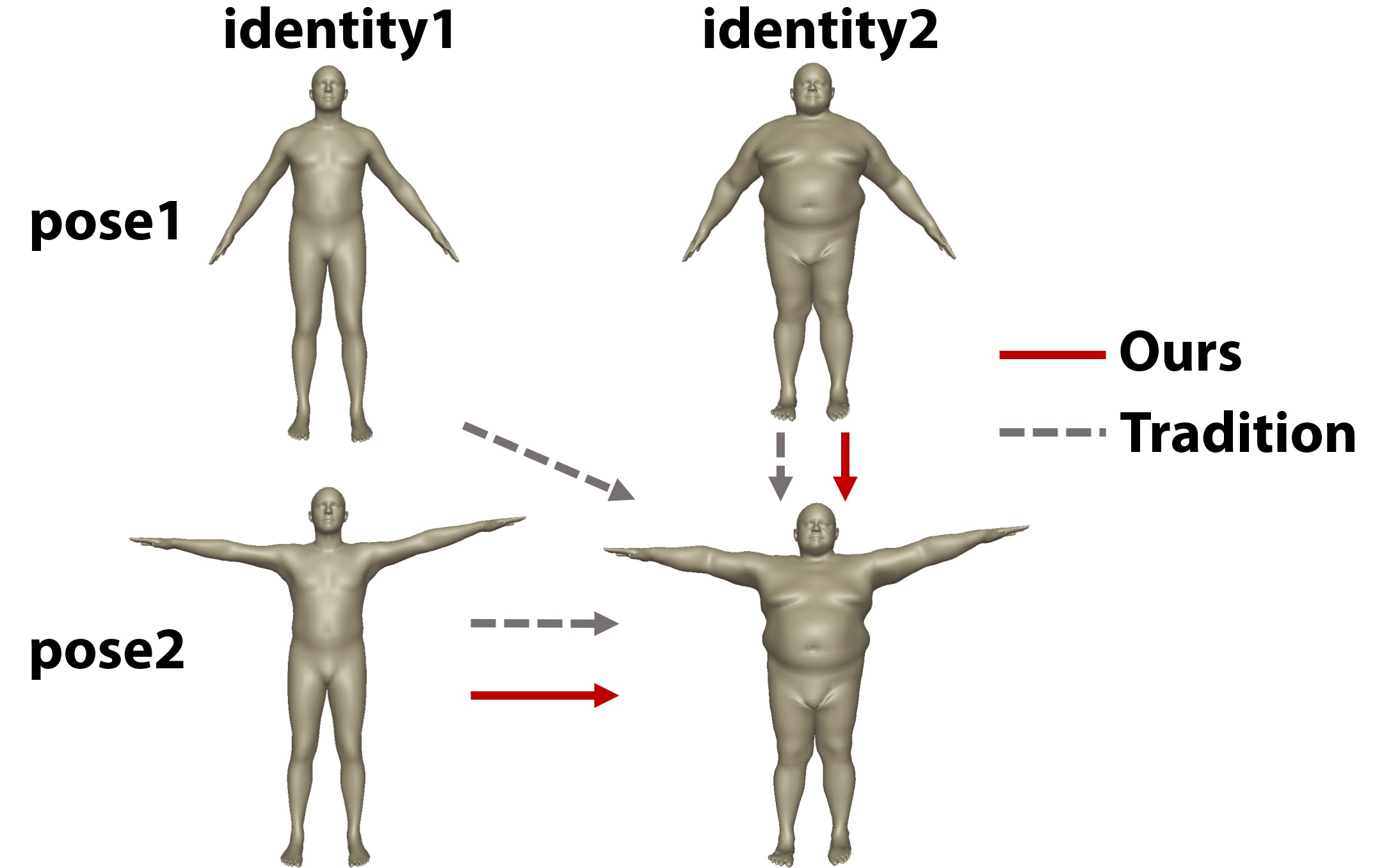}
\end{center}
\caption{\textbf{Pose is transferred from source to the target mesh.} Different from traditional methods, we only require the pose and identity mesh but not any extra input.}
\label{fig:onecol}
\vspace{-1em}
\end{figure}

In this work, we propose a deep learning model for human pose transfer, which transfers the pose from a source mesh to a target identity mesh, as shown in Fig.~\ref{fig:onecol}.
Our model does not rely on any extra auxiliary inputs that implicitly or explicitly build correspondence, and can work for source and target mesh with vertices in random and different order.
These flexibilities make our model very convenient to use in practice and can directly work on identity mesh obtained from arbitrary sources, which however are extremely challenging to be achieved by the  the framework of existing deformation based approaches.
As the output, our model produces a human mesh with the identity from the target mesh and the pose from the source mesh.

Essentially, our key idea is to re-purpose style transfer techniques, which is widely used in image analysis for the deformation transfer problem.
Our model takes the identity information of the target mesh as a ``style'' and transfer it to the source mesh to achieve a pose transfer.
Rather than explicitly learn to deform the target meshes from source meshes, we stack several convolutional layers to gradually encode the pose information from source meshes, and then decode it back to the desired output under the guide of the features learned from the target mesh.
Inspired by the great success of SPADE \cite{huang2017arbitrary} for 2D image, we introduce it to 3D domain to process the point clouds together with PointNet-like network architecture \cite{qi2017pointnet}.

In SPADE \cite{huang2017arbitrary} for 2D image, an affine transformation is learned from each pixel in the target image (``style'') and applied to the instance normalized feature of the corresponding pixel in the source image (``content'').
We propose Spatially Adaptive Instance Normalization (SPAdaIN) to make an analogy between the image pixel and the mesh vertex.
In particular, we first learn a feature vector for each vertex in the source mesh (``pose'') and transform them using an affine transformation learned from a vertex in the target mesh (``identity'').
However, this would not work naively, since the correspondence between source and target is unknown.
To make the feature on the source mesh invariant to the vertex order permutation, the convolutional filters of $1 \times 1$ are utilized to  each individual point and an instance normalization (among all the points) is appended to exchange context global-wise.
The learned features can then be associated with the target mesh vertices in arbitrary permutation for style transfer.
We found this model effectively transfers the unseen identity onto the source pose mesh and produces much more accurate human shape than the state-of-the-art approaches which even requires the additional auxiliary inputs.

\noindent \textbf{Contributions}. 
The contributions of this paper are summarized as follows. To the best of our knowledge, we propose the first end-to-end deep learning model that deforms an identity mesh with the pose from another mesh, even though the identity mesh is unseen and with more fine-grained geometry details. Our model does not require any additional auxiliary mesh or extra knowledge to bridge the huge visual gap between source and target meshes. 
Our model is also convenient to use in practice since the pose and identity mesh can come with different vertices order. Moreover, our model is robust to pose mesh geometry noise. Extensive experiments verify that our model is capable of inferring and transferring the poses from source to target meshes, and the result is invariant to the mesh vertex order between source and target meshes. 


\section{Related Works}

\label{gen_inst}
\paragraph{Deformation transfer.}

Deformation transfer aims to produce a new 3D shape given a pair of
source and target shapes as well as a deformed source shape, making
the target shape do the same deformation (Fig.~\ref{fig:onecol}). 
However, some traditional methods based on skinning skeleton animation \cite{BBW:2011} require additional manual adjustment.
Alternatively, many works leverage affine transformations to generate target shapes \cite{sumner2004deformation,xu2007gradient,yang2018biharmonic}.
Sumner \etal \cite{sumner2004deformation} transfer deformation gradients, but requires corresponding landmarks to handle the differences between shapes. 
Baran \etal \cite{baran2009semantic} assume semantic relationships between the poses of two characters. However, the requirement of semantic similarity pairs limits the usability of this approach.
Ben \etal \cite{ben2009spatial} deform to target shapes with the help of a set of control cages. Chu \etal \cite{chu2010example} proposed to use a few examples to generate natural results.
Even with impressive success, the reliance on auxiliary data makes it difficult to automatically transfer pose for graphics-based methods. To address this, Gao \etal \cite{gao2018automatic} proposed VC-GAN, using cycle consistency to achieve the deformation transfer. But this approach also raises another problem, losing versatility due to over-reliance on training data. Whenever dealing with new identities, it needs to gather training data and retrain the model.

\vspace{-1em}
\paragraph{Deep learning for non-rigid shape representation.}
\cite{Tan_2018_CVPR,litany2018deformable} proposed mesh variational
autoencoder to learn mesh embedding for shape synthesis. However,
they merely use the fully-connected layer which will consume a large amount
of computing resources. \cite{feng2019meshnet} using mesh convolution
to capture the triangle faces feature of 3D mesh. Although their methods
use spatial and structural information, the features represented by
faces are not suitable for our task. Qi \etal proposed
PointNet~\cite{qi2017pointnet} to extract features from unorganized points cloud, but the
missing edge information will result in deformed 3D shape with outliers.
Therefore, we use mesh as the representation of 3D shape, but use
shared weights convolution layers as the network structure of encoder.

\vspace{-1em}
\paragraph{Conditional normalization and style transfer.}
Several conditional normalization methods have been proposed \cite{dumoulin2016learned,NIPS2017_7237,huang2017arbitrary,shi122018conditional}.
At first, they are used in style transfer and then for other vision tasks \cite{huang2018multimodal,miyato2018cgans,perez2017learning,mescheder2018training,zhang2018self,perez2018film} . 
External data is needed in these works. After normalizing the mean and bias of the activation layer, through using these external data they learn the
affine transformation parameters to denormalize the activation layer.
Park \etal~\cite{spade} proposes a similar idea to help with the image synthesis but from a spatial way using the spatially-varying semantic mask. 
This inspires us to apply spatial 3D mesh as external data to generate
our expected mesh. Since the 3d coordinates of the point which 
spatially and naturally are one of the most important representations
of 3D data, the idea of using conditional normalization directly in
the spatial sense is very intuitive and the results from the experiments
demonstrate the effectiveness of this method.

\vspace{-1em}
\paragraph{SPAdaIN vs. other Conditional normalization.} Particularly, we emphasize the difference that: (1) Compare to SPADE~\cite{spade},
we using instance normalization. Since each instance may have different
features to guide the transfer, normalize the activation of the network in
channel-wise is not reasonable. So, we normalize the spatially-variant
parameters instance-wise, which is more suitable for the neural pose
transfer task. (2) Compare to CIN~\cite{dumoulin2016learned}, our normalization parameter vectors
are not selected from a fixed set of identities or pose, the corresponding
parameters $\gamma$ and $\beta$ are adaptively learned, therefore,
their approach cannot adapt to new identities or pose without re-training.
Also, their parameters are aggregated across spatial axes; thus they
may lose some detailed feature in particular spatial positions. (3) AdaIN~\cite{huang2017arbitrary}
is also not suitable for pose transfer. Though AdaIN can handle
arbitrary new identities or pose as guidance, there are no learnable
parameters in AdaIN. Due to the lake of learnable parameters, when
adopting AdaIN as normalization, the network will tend to imitate
the shape of $\mM$ rather than use it as a condition to produce new posture.


\begin{figure}[t]
\begin{center}
\includegraphics[width=\columnwidth]{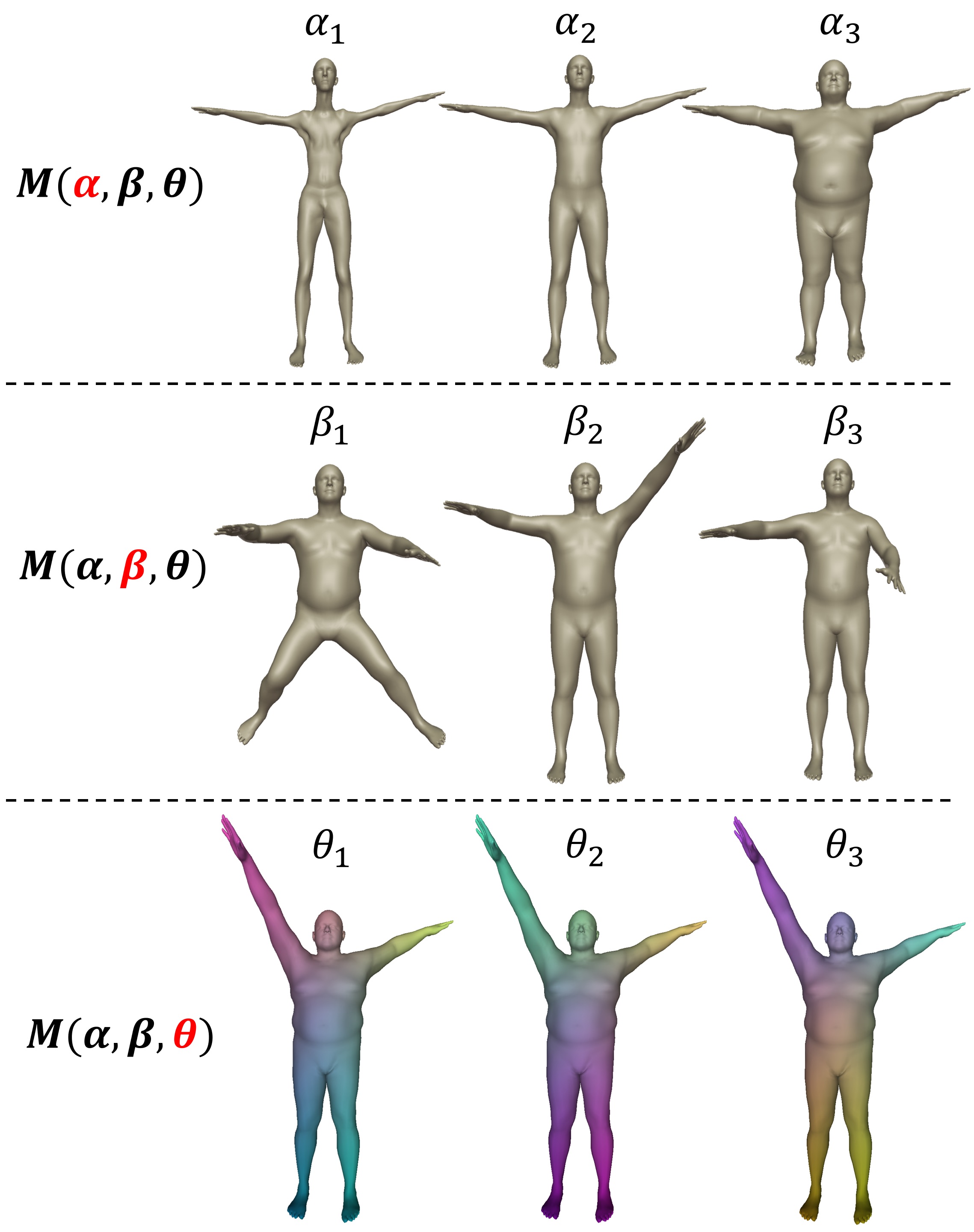}
\end{center}
\vspace{-5pt}
\caption{\textbf{Influence of different parameters on 3D human mesh model.} Each row represents the change of the mesh when changing one parameter from $\alpha,\beta,\theta$. $\alpha$ controls the mesh identity, $\beta$ controls the mesh pose and $\theta$ indicates the vertices order. The mesh color of last row encodes the mesh vertex index.}
\label{fig:para}
\vspace{-0.6cm}
\end{figure}
\vspace{-0.2cm}
\section{Methods}
\begin{figure*}[t]
\begin{center}
\includegraphics[width=1\linewidth]{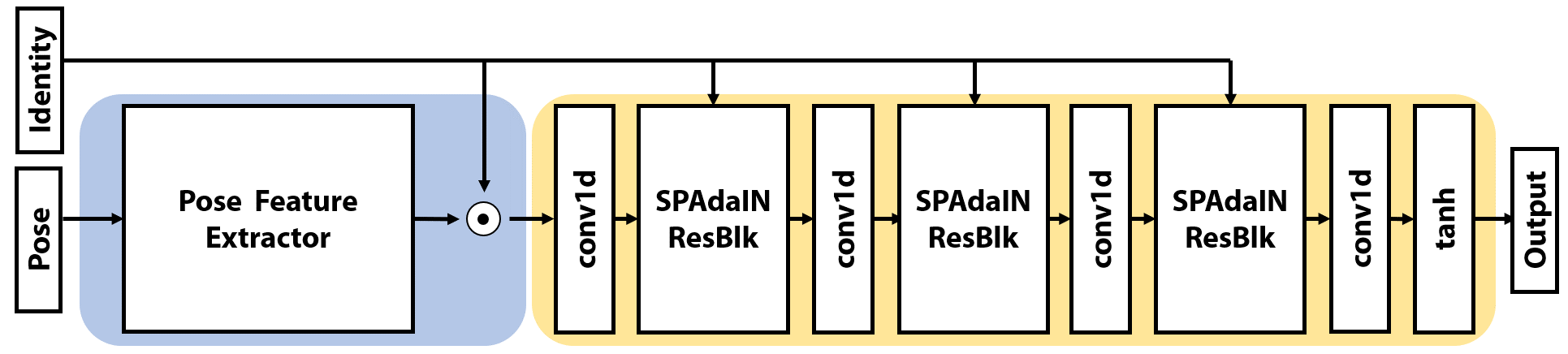}
\end{center}
   \caption{\textbf{Network Architecture.} The blue part is permutation
invariant encoder, and the yellow part is SPAdaIN guided decoder.
Given $\mM_{id}$ and $\mM_{pose}$ as input, we produce mesh transferred
to new posture. The symbol $\odot$ denotes the  operation of concatenation.}
\label{fig:pipeline}
\end{figure*}

In this section, we introduce our deep learning model for human pose transfer (Fig.~\ref{fig:pipeline}). 
Our model is highly inspired by image style transfer.
Taking the source mesh carrying the pose, our model produces a feature for each vertex encoding both local details and global context.
The per-vertex features are then concatenated with the vertex locations in the target mesh providing identity, which is fed into the style transfer decoder consists of SPAdaIn ResBlocks.
Throughout the decoder, each feature produces one vertex in the output mesh under the guide of a vertex from the target mesh.
The final output mesh inherits the pose from the source mesh and the identity from the target.
The mesh vertex order is consistent with the identity mesh.

\begin{figure*}[t]
\begin{center}
\includegraphics[width=1\linewidth]{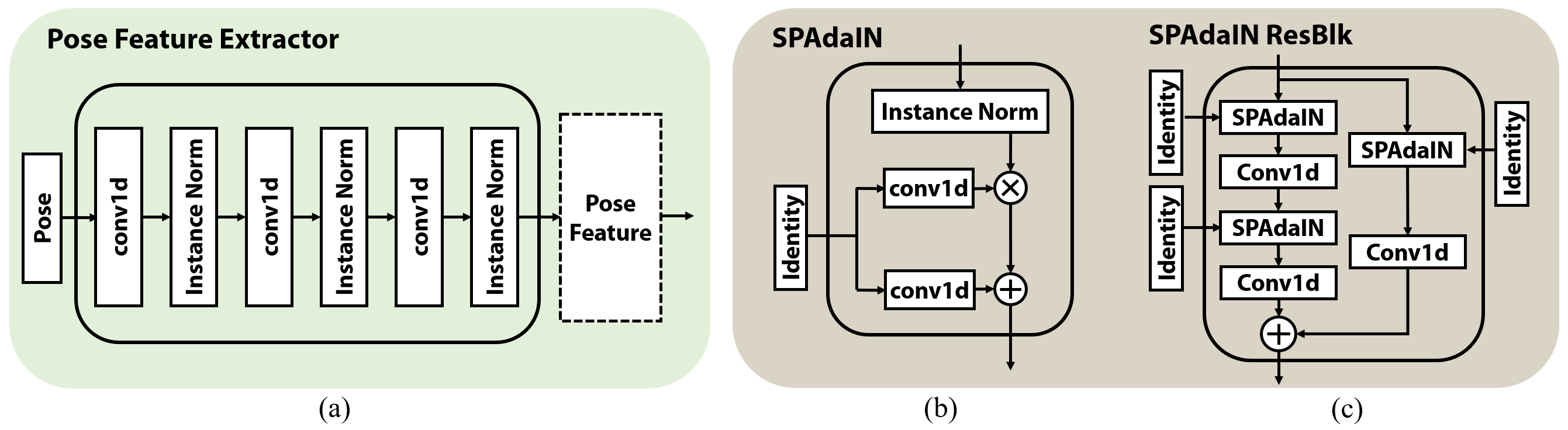}
\end{center}
   \caption{\textbf{Detailed Network Component Architecture.} (a) Architecture of Pose Feature Extractor, (b) Architecture of SPAdaIN and (c) Architecture of SPAdaIN ResBlock.}
\label{fig:component}
\vspace{-2em}
\end{figure*}

\subsection{Problem Definition}
We represent 3D human mesh by $\mM(\alpha, \beta, \theta)$. 
As shown in Fig.~\ref{fig:para}, $\alpha$ denotes the parameters of mesh identity, which controls the mesh shape, $\beta$ represents different human posture, $\theta$ indicates the vertices order.
Given two meshes $\mM_{id}=\mM(\alpha_1, \beta_1, \theta_1)$ and $\mM_{pose}=\mM(\alpha_2, \beta_2, \theta_2)$, our goal is to transfer the pose to the identity mesh by producing output mesh $\mM_{output}=\widehat{\mM}(\alpha_1, \beta_2, \theta_1)$.

\subsection{Permutation Invariant Pose Feature Extractor}
We first introduce our pose feature extractor $\mathbf{E}$. The encoder aims to extract the feature $\mF_{pose}$ for the orderless input mesh vertices.
The encoder $\mathbf{E}$ takes $\mM_{pose}$ vertices coordinates through pose feature extractor as illustrated in Fig.~\ref{fig:component} (a). The pose feature extractor consists of 3 stacked $1\times 1$ convolution and InstanceNorm layers, all activation functions applied for convolution layers are ReLU. 
Then the encoder concatenates pose features with the vertices coordinates of template identity mesh $\mM_{id}$ to produce latent embedding $\mZ=\mF_{pose}\odot \mM_{id}$ eventually ($\odot$ denotes concatenation).
One architecture choice needs to be discussed is why $\mF_{pose}$ are tensors rather than global vectors. Since the vertex orders of different training data are not consistent, and normalization is essential to aggregate the global context, InstanceNorm (IN) is the only choice to normalize the features. However, if $\mathbf{E}$ encodes pose feature as a global vector and then attaching it to $\mM_{id}$, calculating IN will lead the pose features to be normalized to zero. So we prefer to learn the pose feature with  the same spatial size as $\mM_{id}$. In principle, this  will allow the whole pipeline to preserve spatial information and be free from the requirement of point-wise correspondence between $\mM_{id}$ and $\mM_{pose}$.

\vspace{-1em}
\subsection{Style Transfer Decoder}
In this section, we introduce our novel condition normalization layer SPAdaIN first. Then we describe the decoder architecture build upon SPAdaIN ResBlock. 

\noindent \textbf{SPAdaIN.} Extending previous style transfer works \cite{dumoulin2016learned,huang2017arbitrary,spade}, we propose spatially conditional normalization to generate the 3D human shape applied to pose transfer tasks while keeping the identity of meshes. In particular, SPAdaIN is a generalization
of \cite{huang2017arbitrary,spade} to deal with points. Similar to IN,
the activation is normalized across the spatial dimensions independently
for each channel and instance, and then modulated with learned
scale $\gamma$ and bias $\beta$. Note that, here we assume that in the $i$-th
layer, $\mM$ is the 3D model providing identity, $V^{i}$
is the number of 3D shape vertices in this layer, $C^{i}$ is the number
of feature channel, $N$ denotes the batch size, and $h$ is the activation
value of network (the footnote indicate specific index where $n\in N,c\in C^{i},v\in V^{i}$).
The value normalized by SPAdaIN can be computed as follows,
\begin{equation}
\mu_{n,c}^{i}=\frac{1}{V^{i}}\sum_{v}h_{n,c,v}^{i}
\end{equation}
\begin{equation}
\sigma_{n,c}^{i}=\sqrt{\dfrac{1}{V^{i}}\sum_{v}\left(h_{n,c,v}^{i}-\mu_{n,c}^{i}\right)^{2}+\varepsilon}
\end{equation}
\begin{equation}
\mathrm{SPAdaIN}(h,\mM)=\gamma_{v}^{i}(\mM)\left(\frac{h_{n,c,v}^{i}-\mu_{n,c}^{i}}{\sigma_{n,c}^{i}}\right)+\beta_{v}^{i}(\mM)
\end{equation}

\noindent where $\gamma$ and $\beta$ are learnable affine parameters, $\varepsilon=1e-5$ for numerical stability.
The detailed SPAdaIN module structure is shown in Fig.~\ref{fig:component} (b). In SPAdaIN the external data $\mM_{id}$ is fed into 2 different $1\times 1$ convolution layers to produce the modulation parameters $\gamma$ and $\beta$. The parameters are multiplied and added to the normalized feature.

\noindent\textbf{Decoder.} 
The decoder architecture we employed is inspired by the style transfer task.
We first feed the latent embedding $\mZ$ into the decoder, consisting of multiple SPAdaIN ResBlocks. As shown in Fig.~\ref{fig:pipeline}, the overall architecture has $3$ SPAdaIN ResBlocks.
Fig.~\ref{fig:component} (c) illustrates the detail of SPAdaIN ResBlock architecture. Each SPAdaIN ResBlock consists of SPAdaIN blocks followed by a $1\times 1$ convolution layer and ReLU activation function, and 3 identical units are organized in the form of residual block \cite{he2016deep}. The output of this operation is then fed to a tanh layer, generating the final output $\mM_{output}$.

\subsection{Loss function}
To efficiently train our network, we introduce and define the loss function $\mathcal{L}$ as follows, 
\begin{equation}
\mathcal{L}=\mathcal{L}_{rec}+\lambda_{edg}\cdot\mathcal{L}_{edg}
\end{equation}
where $\lambda_{edg}$ is coefficients of edge regularization. 
\vspace{-1em}
\paragraph{Reconstruction Loss.}
The loss aims to regress the vertices close to its correct
position.We pre-process the ground truth with the same vertices number
as template identity model and train the network using the supervision
of point-wise L2 distance between the mesh predicted by our model $\widehat{\mM}(\alpha_1, \beta_2,\theta_1)$ and the ground truth mesh $\mM(\alpha_1, \beta_2,\theta_1)$.

\begin{equation}
\mathcal{L}_{rec}=||\widehat{\mM}(\alpha_1, \beta_2,\theta_1)-\mM(\alpha_1, \beta_2,\theta_1)||_{2}^{2}
\end{equation}
\vspace{-2em}
\paragraph{Edge Length Regularization.}
Directly regress vertices position will not guarantee that the transferred
of avoiding producing the over-length edges, since we tend to make the generated 
model has smooth surface. To address this problem, we further propose edge length regularization penalizing the long edges. Specifically, this regularization enforces the output mesh surface
to be tight, resulting in a smooth surface. Inspired by \cite{groueix20183d},
let $\mathcal{N}(p)$ be the neighbor of vertex $p$, the edge length
regularization can be defined as follows,

\begin{equation}
\mathcal{L}_{edg}=\sum_{p}\sum_{v\in\mathcal{N}(p)}||p-v||_{2}^{2}
\end{equation}
\vspace{-2em}
\section{Experiment}
\subsection{Experimental setup}
\paragraph{Datasets.}
We use SMPL model \cite{loper2015smpl} to generate training and test data by randomly sampling the parameter space.
To create training data, we generate meshes of 16 identities with 400 poses, and randomly pick two as a pair for training.
The ground truth is obtained by running SMPL model\cite{loper2015smpl} with the desired shape and pose parameters from two meshes respectively.
In order to be invariance to  the vertex order, the mesh vertices are shuffled randomly before feeding into the network. Accordingly,
the ground truth mesh is shuffled in the same manner as the identity mesh such that they are point-wise aligned to its corresponding input mesh.

In the test step, we evaluate our model for transferring the seen and unseen poses to new identities.
To do so, we create 14 new identities that are not in the training set.
We use these new identities to form 72 pairs with randomly selected training pose, and 72 pairs with newly created poses.
To further test how our model generalizes, we employ the meshes from FAUST \cite{Bogo_2014_CVPR} and MG-dataset\cite{bhatnagar2019mgn}.
These meshes are not strictly consistent with SMPL but with more fine-grained geometry details and more realistic.

For all input meshes, we shift them to the center and scale them to the unit sphere, our method is robust against the global scale.

\vspace{-1em}
\paragraph{Implementation details.}
The hyper-parameters to train our network are as follows.
We use Adam optimizer with the learning rate as $5e-5$.
The $\lambda_{edg}$ in the loss function is set as $5e-4$.
The model is trained for 200 epochs with batch size equalling to 8 on a single GTX 1080Ti GPU.
Please refer to the supplementary material for more detailed network architecture.

\vspace{-1em}
\paragraph{Evaluation Metrics.}
Since the output mesh is point-wise aligned with the ground truth,
we use Point-wise Mesh Euclidean Distance (PMD) as our evaluation
metrics. Specifically, 
\begin{equation}
\mathbf{PMD}=\frac{1}{|V|}\sum_{v}||P_{v}-Q_{v}||_{2}^{2}
\end{equation}
where we have mesh vertices $P_v \in \widehat{\mM}(\alpha_1, \beta_2,\theta_1)$ and $Q_v \in \mM(\alpha_1, \beta_2,\theta_1)$.

\begin{figure*}[t]
\begin{center}
\includegraphics[width=1\linewidth]{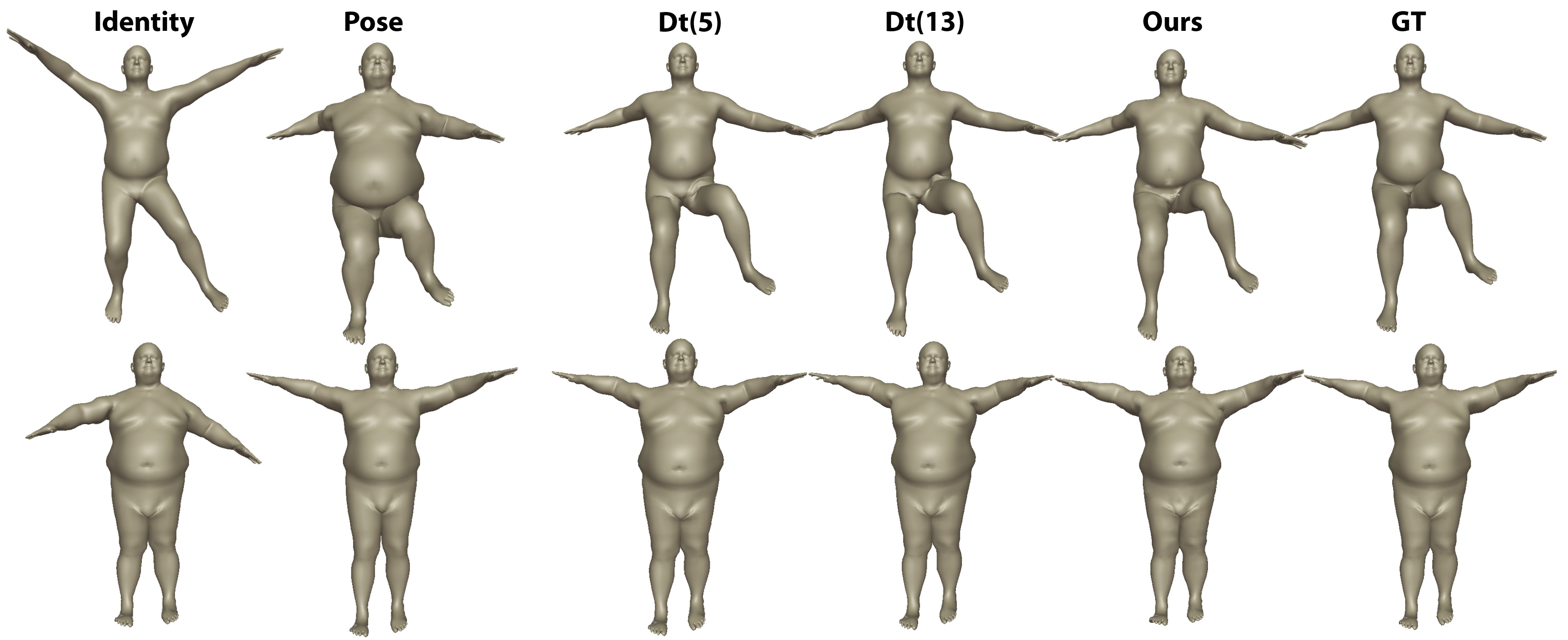}
\end{center}
   \caption{\textbf{Qualitative comparison of seen pose.} From left to right, we show in each row: input identity mesh, input pose mesh, the results of DT~\cite{sumner2004deformation} using 5 control points and 13 control points respectively, our results and the ground truth. Our predictions have more natural joints movement. }
\label{fig:compare_sup}
\end{figure*}
\begin{figure*}[h]
\begin{center}
\includegraphics[width=1\linewidth]{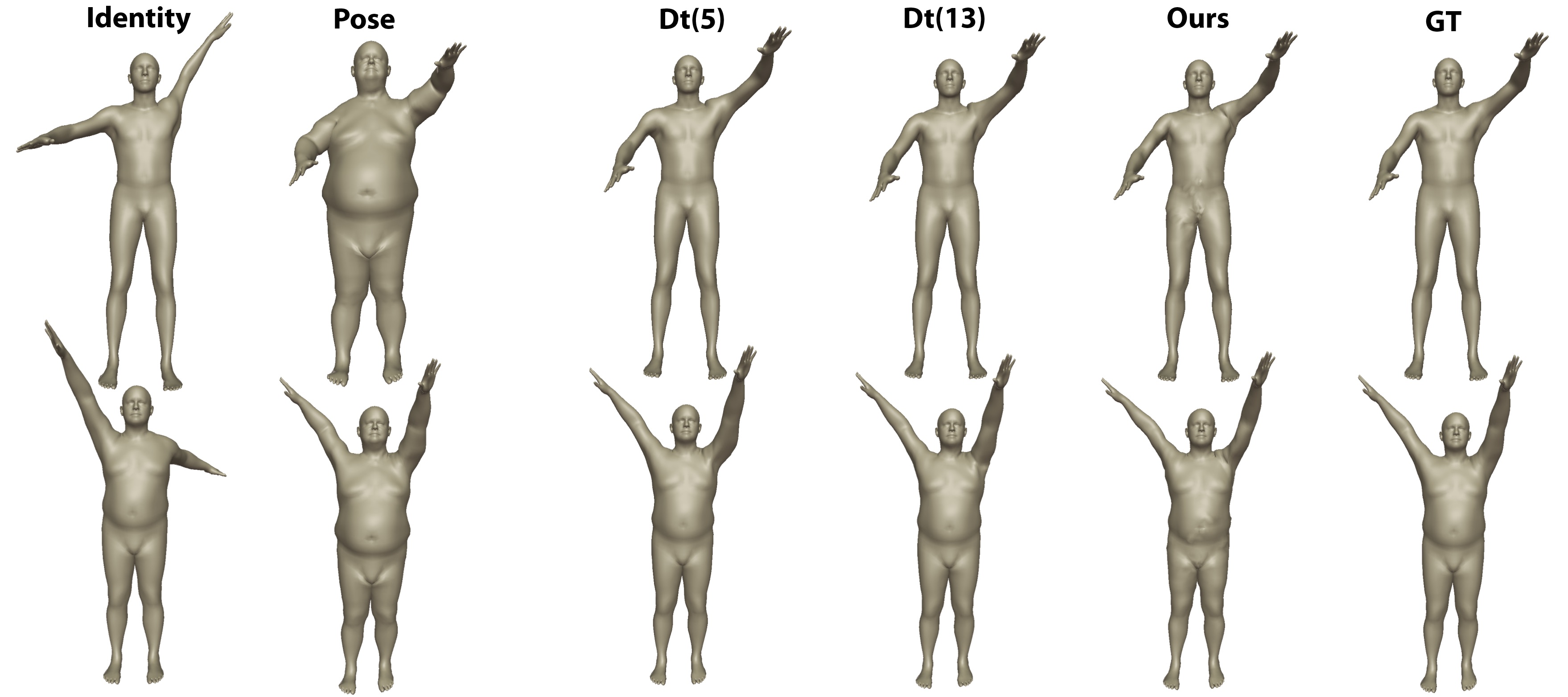}
\end{center}
\caption{\textbf{Qualitative comparison of unseen pose.} Both $\mM_{id}$ and $\mM_{pose}$ are unseen. From left to right, we show in each row: input identity mesh, input pose mesh, the results of DT~\cite{sumner2004deformation} using 5 control points and 13 control points respectively, our results and the ground truth. Our predictions are more natural at joints.}
\vspace{-1em}
\label{fig:compare_unsup}
\end{figure*}

\subsection{Comparison to Deformation Transfer}

In this section, we compare to deformation transfer baselines and show both qualitative and quantitative results.
To the best of our knowledge, there are no learning-based methods for
deformation transfer designed for new identities yet. One of the most effective methods is deformation transfer (DT) \cite{sumner2004deformation}, which however, has to rely on the additional control points and a third mesh, as the auxiliary input. To this end, 
we provide DT the third mesh and run it with 5 and 13 control points.
The qualitative results are shown in Fig.~\ref{fig:compare_sup} and Fig.~\ref{fig:compare_unsup} and quantitative results are shown in Tab. \ref{tab:compare}.
As can be seen,  our model learns  on seen poses to transfer poses effectively to new identities in the testing set, while the PMD of our method is significantly lower than that of DT which even has the additional inputs. This greatly validates the effectiveness of our model in learning to deform meshes.
Furthermore, for those poses that have never been seen during the training stage, our model demonstrates very good generalization ability and still produces reasonable good results as shown in Fig.~\ref{fig:compare_unsup}. 
Note that DT is not a learning-based approach such that it has quite similar performance over the training and testing set.

\begin{figure}[h]
\begin{center}
\includegraphics[width=\columnwidth]{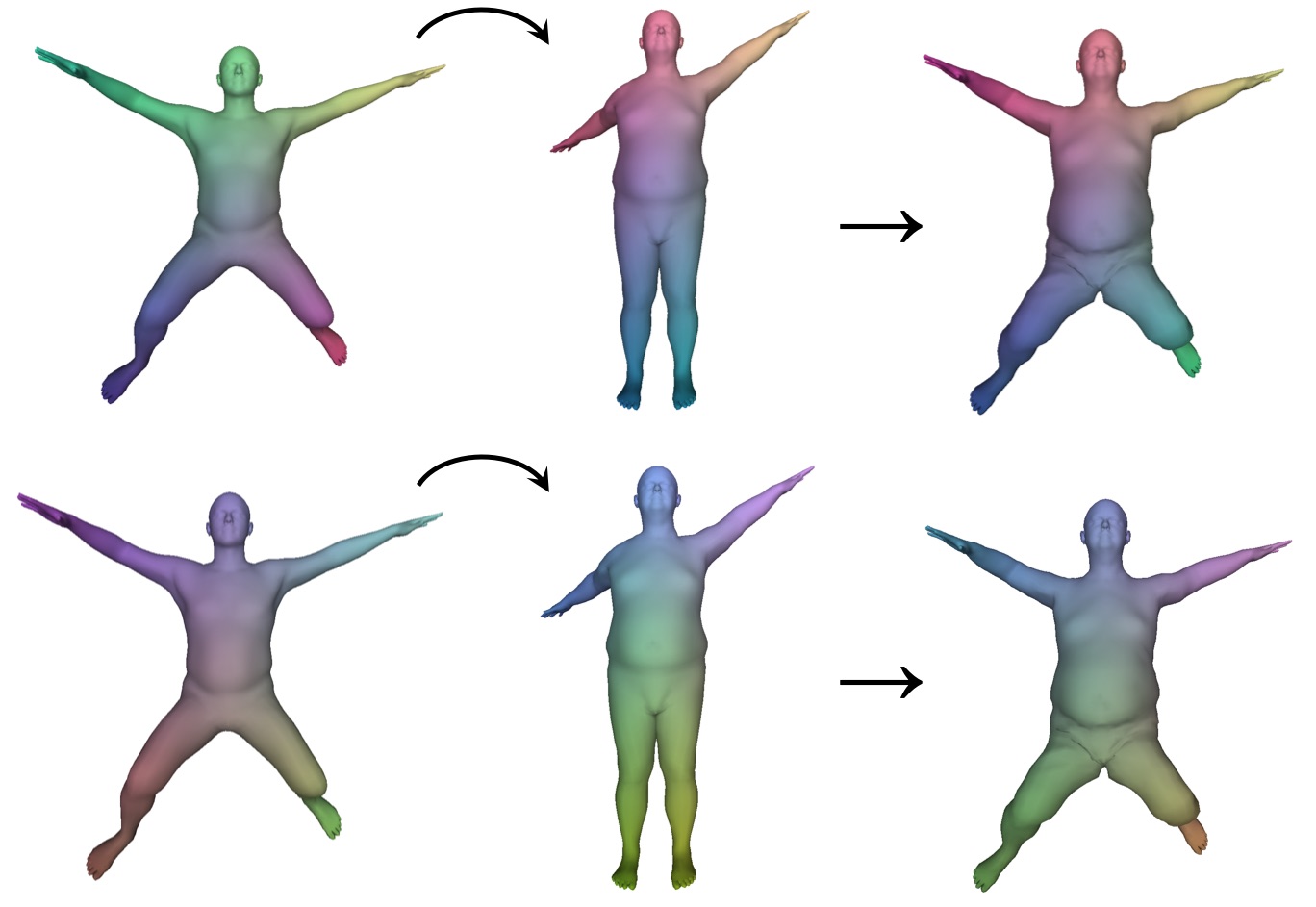}
\end{center}
\vspace{-5pt}
\caption{\textbf{Visualization of the vertex index color encoding.} We show two pairs of input meshes with different vertices orders and the predict results. From left to right, $\mM_{pose}$, $\mM_{id}$ and $\mM_{output}$. The order of our pose transfer results are consistent with the identity mesh.}
\vspace{-1em}
\label{fig:dif_order}
\end{figure}

To demonstrate that our model is invariant to vertex permutation of meshes, we further run our model on the same pair of meshes with the identity mesh shuffled in different orders.
Fig.~\ref{fig:dif_order} shows the input and output meshes with color encoding the vertex index. 
As can be seen, our model can produce similar output meshes with the input identity meshes in different shuffles. This shows that the output vertex order is maintained the same as the identity meshes.

\begin{table}[tbh]
\caption{\textbf{Quantitative comparison of average PMD.} }
\label{tab:compare}

\centering{}%
\begin{tabu} to \columnwidth {X[2]X[1.5c]X[1.7c]X[c]}
\toprule
\multirow{3}{*}{Pose Type} &
\multicolumn{3}{c}{PMD~$\downarrow$~($\times 10^{-4}$)}\\
\tabuphantomline
\cmidrule(lr){2-4}
& DT(5) \cite{sumner2004deformation} & DT(13) \cite{sumner2004deformation} & Ours \tabularnewline
\midrule 
\midrule 
seen-pose  & 7.3  & 7.7 & 1.1 \tabularnewline
unseen-pose  & 7.2  & 6.7 & 9.3 \tabularnewline
\bottomrule
\end{tabu}
\end{table}

\vspace{-20pt}


\begin{figure}[t]
\begin{center}
\begin{tabular}{c}

\includegraphics[width=1\columnwidth]{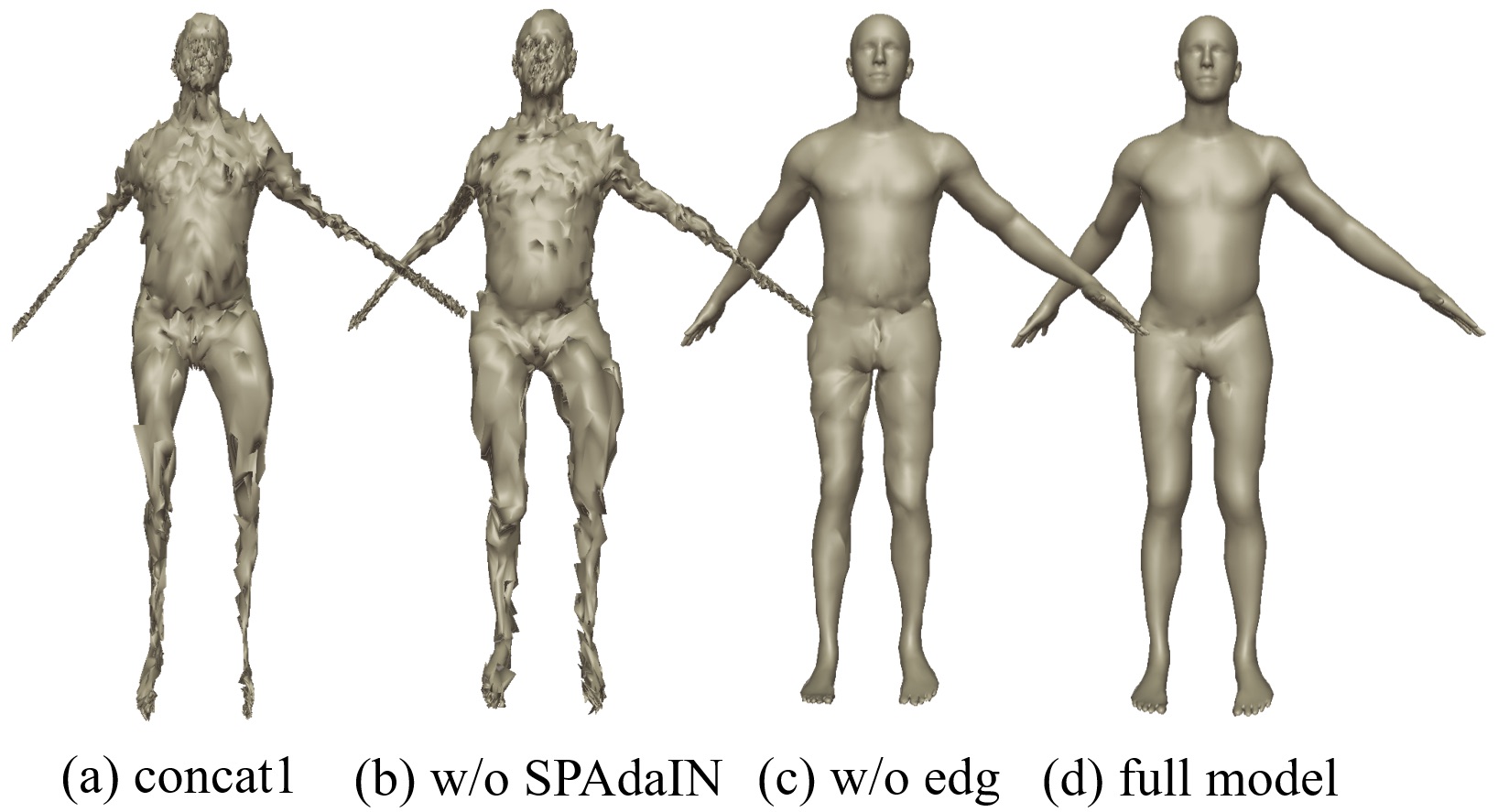}
\end{tabular}
\end{center}
\caption{\textbf{Qualitative ablation study results.} We show the generation results of (a) our naive baseline concat1, (b) our model without SPAdaIN modules, (c) our model without edge regularization and (d) our full model. As we can see, SPAdaIN is very helpful in learning the pose transfer and edge loss can help to generate smoother results.}
\label{fig:ablation}
\vspace{-10pt}
\end{figure}

\begin{figure}[t]
\begin{center}
\begin{tabular}{c}
\vspace{-5pt}
\includegraphics[width=0.8\columnwidth]{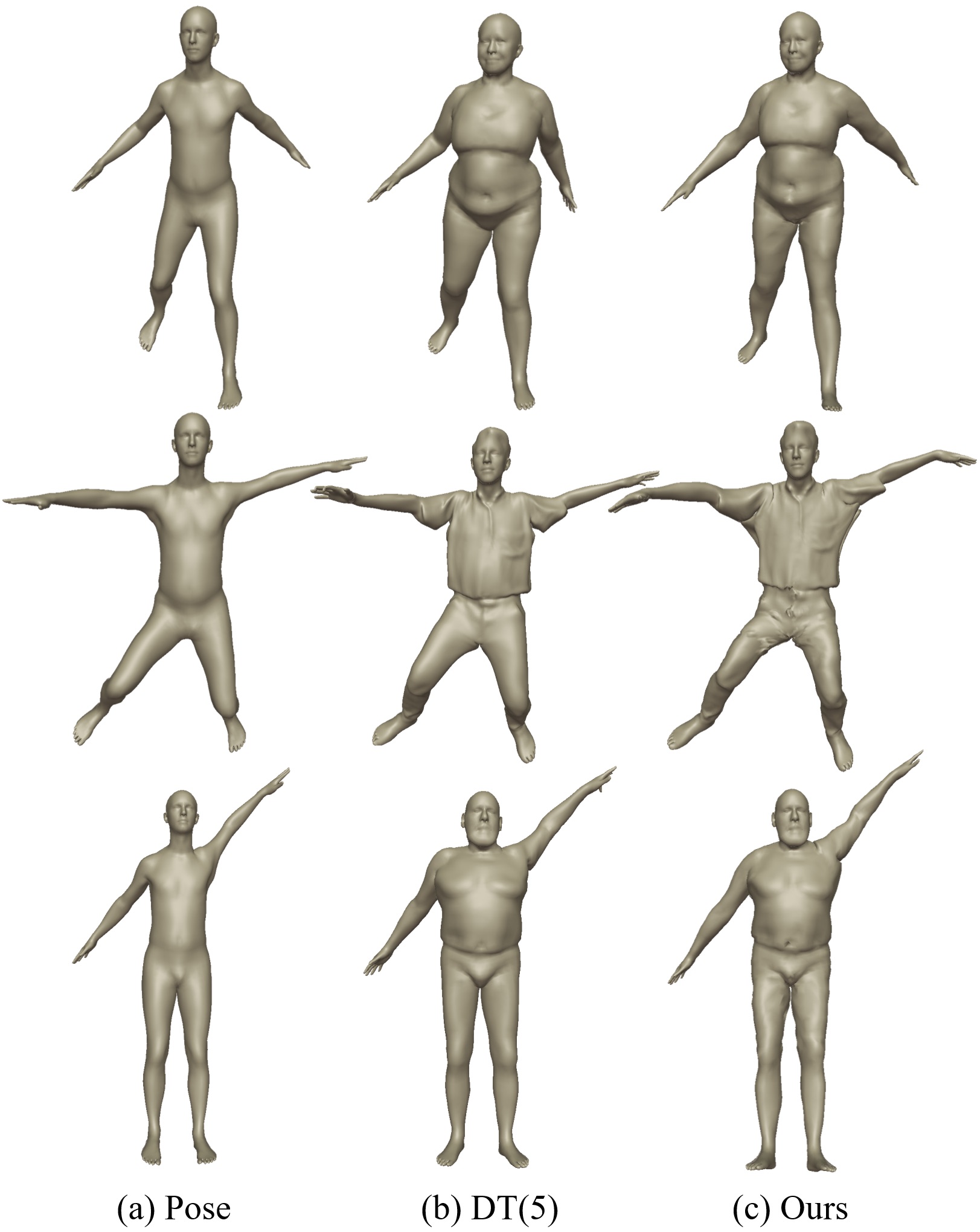}
\end{tabular}
\end{center}
\caption{\textbf{Qualitative comparisons of Non-SMPL based identity.} Comparison of using FAUST \cite{Bogo_2014_CVPR} and MG-dataset\cite{bhatnagar2019mgn} as identity mesh respectively.} 
\label{fig:compare_faust_cloth}
\vspace{-2em}
\end{figure}

\subsection{Ablation Study}
In this section, we verify the effectiveness of the key components of our model by some ablation study.

We start from a naive network architecture, where the decoder only consists of several 1-dimensional convolutional filters (conv1d).
We then sequentially add ResBlock and SPAdaIN to the network.
We name these two naive methods concat1 and w/o SPAdaIN.
The quantitative evaluations are shown in Tab.~\ref{tab:ablation}, and some examples can be found in Fig. \ref{fig:ablation}.
As can be seen,  naive conv1d (concat1) does not perform well, and the surface details are added back gradually when adding more components to the network. Particularly, SPAdaIN is very helpful in learning the pose transfer, which reduces the error from $8.3$ to $1.1$ on seen poses and from $13.7$ to $9.3$ on unseen poses.
This means the style transfer network can effectively transfer the identity as a style onto the target mesh.

We also evaluate the impact of edge regularization loss on the model performance. As compared in Tab. \ref{tab:ablation}, edge regularization loss consistency reduces the PMD on the testing dataset of both seen and unseen poses. From Fig. \ref{fig:ablation}, the results are smoother if trained with the edge loss,  compared to those without using edge regularization loss.

\begin{table}[tbh]
\caption{\textbf{Quantitative ablation study for seen and unseen pose.}
We show the metrics of PMD with a naive basline (concat1), SPAdaIN and edge regularization disabled respectively, full denotes our full model.}
\label{tab:ablation}
\centering{}%
\begin{tabu} to \columnwidth {X[2.3]X[1c]X[2.5c]X[1.5c]X[0.6c]}
\toprule
\multirow{3}{*}{Pose Source} &
\multicolumn{4}{c}{PMD~$\downarrow$~($\times 10^{-4}$)}\\
\tabuphantomline
\cmidrule(lr){2-5}
& concat1 & w/o SPAdaIN & w/o edg  & full \tabularnewline
\midrule 
\midrule 
seen-pose & 12.1 & 8.3 & 1.2 & 1.1\tabularnewline
unseen-pose &16.9 & 13.7 & 10.1 & 9.3 \tabularnewline
\bottomrule
\end{tabu}
\vspace{-1em} 
\end{table}
\begin{figure}[h!]
\begin{center}
\begin{tabular}{c}
\includegraphics[width=1.\columnwidth]{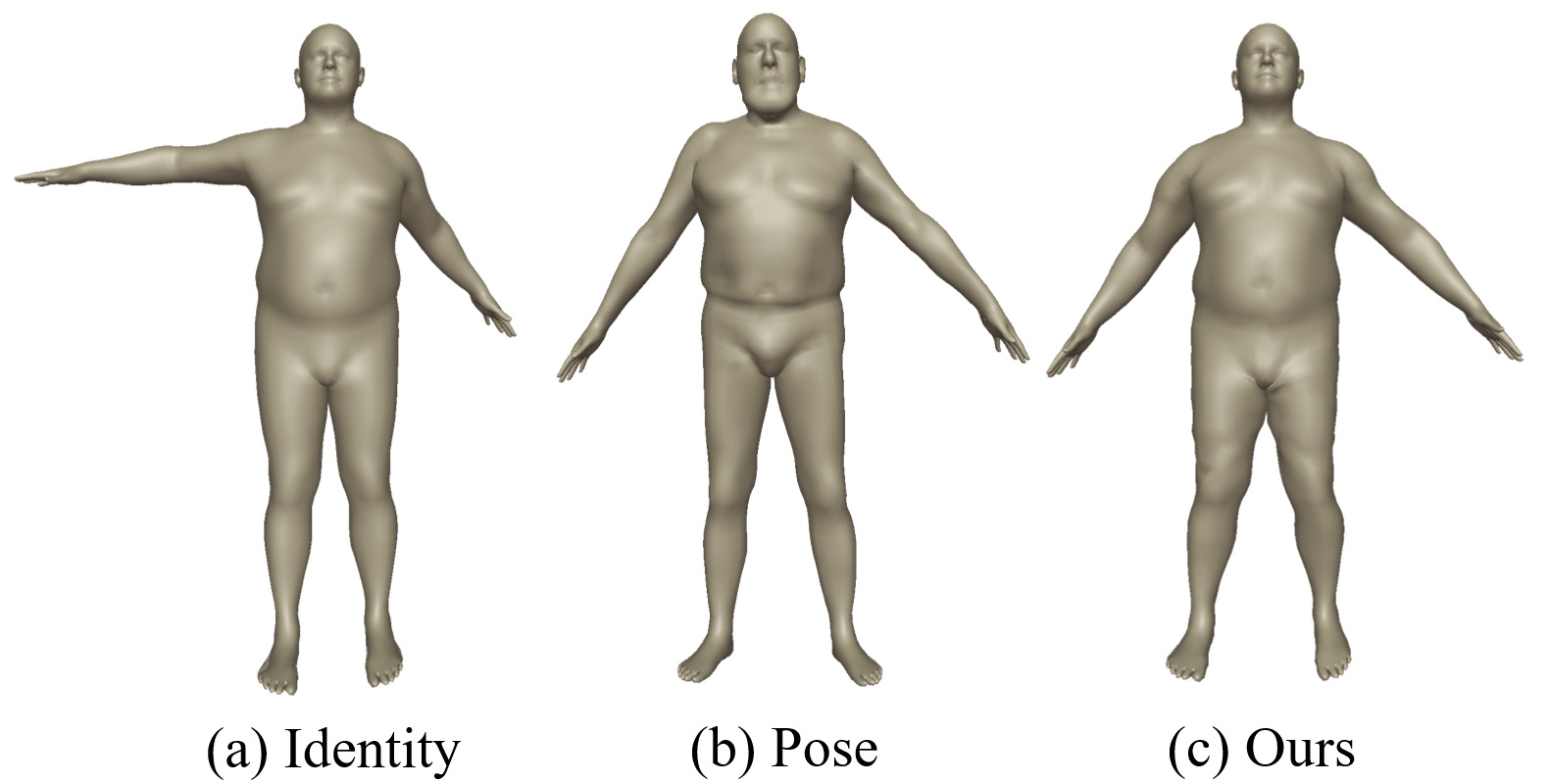}
\tabularnewline
\end{tabular}
\end{center}
\caption{\textbf{Qualitative example of Non-SMPL based pose.} We show results using the mesh from FAUST~\cite{Bogo_2014_CVPR} as pose mesh. Our system has the ability to transfer pose from Non-SMPL based mesh.}
\vspace{-0.2cm}
\label{fig:faust_as_pose}
\end{figure}
\vspace{-1em}
\subsection{Generalization Capability}
In this section, we investigate the generalization capability of our method from the cross source data and robustness. Specifically, we test our model with non-SMPL based identity and pose meshes.
It is worth noting that the training data created by SMPL are highly constrained and lack of geometry details. Our deep learning model can handle the details beyond SMPL capacity decently.
\vspace{-1em}
\paragraph{Non-SMPL based identity}
We first test how our model performs with a human mesh that is not strictly an SMPL model.
To do so, we take meshes from FAUST \cite{Bogo_2014_CVPR} and MG-dataset  \cite{bhatnagar2019mgn} which include dressed human meshes as the identity meshes.
The model we get through SMPL training dataset does not require the order of the mesh vertex points or the number of the points as input, but it has to set the same  number of points of pose mesh, as that of the identity mesh points.
SMPL meshes have $6890$ points each and FAUST  has the same number of points as SMPL. 
For MG-dataset \cite{bhatnagar2019mgn} which has meshes with 27554 each, we adopt the interpolation to automatically increase the number of points of pose mesh and this is a very simple process.
In Fig.~\ref{fig:compare_faust_cloth}, we can see that, even with the identity mesh that is not from SMPL, our model still produces the correct pose while maintains the geometry details that are not encoded by SMPL, \eg, beard and the cloth.
On the other hand, DT sometimes produces more obvious artifacts near fingers.

\begin{figure}[t]
\begin{center}
\begin{tabular}{c}
\includegraphics[width=\columnwidth]{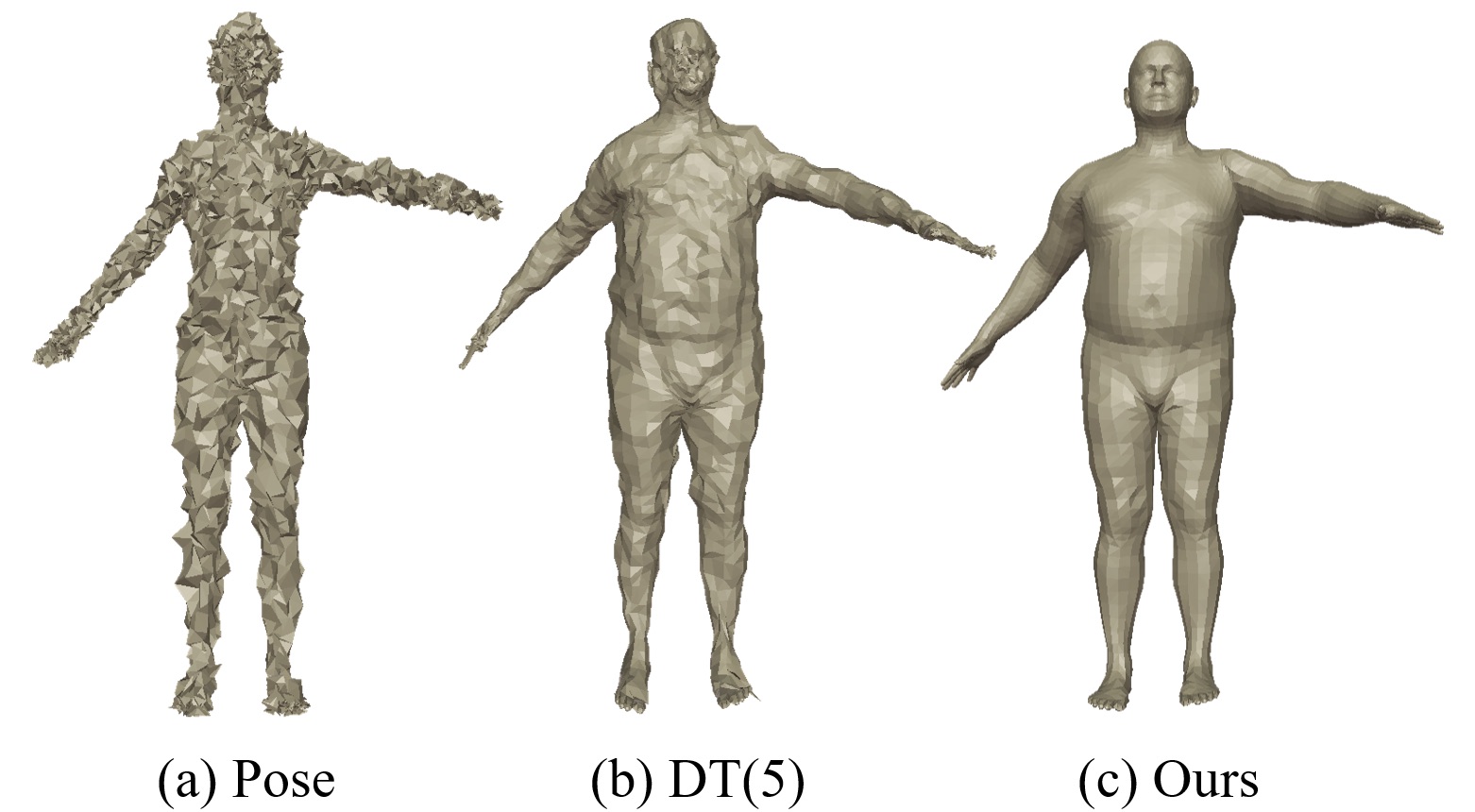} \tabularnewline
\end{tabular}
\end{center}
\caption{\textbf{Robustness to noise.} When using pose mesh (a) with noise, our method (c) still performs very well, however, DT~\cite{sumner2004deformation} (b) may maintain undesirable geometry noise.}
\vspace{-1em}
\label{fig:noise}
\end{figure}

\vspace{-1em}
\paragraph{Non-SMPL based pose.}
We then test our system with a non-SMPL based source mesh which provides the pose.
We give examples using a mesh from FAUST~\cite{Bogo_2014_CVPR} as pose mesh in Fig.~\ref{fig:faust_as_pose}.
As shown in Fig.~\ref{fig:faust_as_pose}, our model still managed to produce reasonably good results.

\vspace{-1em}
\paragraph{Robustness to noise}
Lastly, we test the model robustness against noise in the pose mesh.
We manually add noise to the pose mesh by adding random perturbations to point coordinates, since there may be some noise during application sometimes. 
Surprisingly, as shown in Fig.~\ref{fig:noise}, our method is still doing well.

\vspace{-0.5em}
\section{Conclusion}
In this paper, we propose an efficient deep learning based architecture to efficiently transfer the pose from source meshes to target meshes. The whole network is designed as generalizing the style transfer in the image domain to deal with points.
The novel component -- SPAdaIN is thus introduced to implement our idea.
Strikingly, we empirically validate and show that our network has the potential ability in generalizing to transfer poses to unseen meshes and being invariant to different vertex orders of source and target meshes.
Comparing with the other methods, we show that our model can work well in transferring poses in noisy conditions and in handing arbitrary vertex permutation, most importantly without relying on the additional input from auxiliary meshes or extra knowledge as previous works.

\vspace{-0.5em}
\section*{Acknowledgement}
This work was supported in part by NSFC Projects (U1611461) , 
Science and Technology Commission of Shanghai Municipality Projects (19511120700, 19ZR1471800), Shanghai Municipal Science and Technology Major Project (2018SHZDZX01), and  Shanghai Research and Innovation Functional Program (17DZ2260900).

\clearpage
\renewcommand\thesection{\Alph{section}}

\setcounter{section}{0}
\begin{widetext}
\begin{center}
\textbf{\Large Supplementary Materials}
\end{center}
\end{widetext}

We provide details about network architecture, implementation details, comparison to more baselines, model analysis, and more results on various datasets.

\section{Network Architecture}
The network architecture is shown in Tab. \ref{tab::full_network}, where $N$ is the batch size and $V$ is the number of vertices.
Our network consists of two main parts - the pose feature extractor (1-9) and the style transfer decoder for pose transfer (10-17).
Both components are composed of $1 \times 1$ convolution and instance normalization.
The detailed architecture of SPAdaIN Resnet Block and SPAdaIN unit are given in Tab.~\ref{tab::spadain_resblock} and Tab.~\ref{tab::spadain}.

Different from most of other work that uses batch normalization, we use instance normalization.
Specifically, we consider our input 3D mesh $\mM \in \mathbb{R}^{N\times 3\times V}$ as a tensor and apply normalization individually for each training instance along the spatial dimension $V$. Furthermore, as mentioned in Sec 3.3 of the main submission, we learn the parameters $\gamma \in \mathbb{R}^{N\times C\times V}$ and $\beta \in \mathbb{R}^{N\times C\times V}$ of Instance Norm which keep the spatial information.

\begin{table}[h]
\centering
\begin{adjustbox}{max width=\columnwidth}
\begin{tabular}{|c|c|c|c|}
\hline
Index & Inputs& Operation       & Output shape \\ \hline
(1)   & Input & Identity Mesh   & N$\times$3$\times$V        \\ 
(2)   & Input & Input Features  & N$\times$C$\times$V        \\ 
(3)   & (1), (2)             & SPAdaIN 1 (C=C)         & N$\times$C$\times$V        \\ 
(4)   & (3)   & conv1d(C$\rightarrow$C, $1\times1$), Relu & N$\times$C$\times$V        \\ 
(5)   & (1), (4)             & SPAdaIN 2 (C=C)         & N$\times$C$\times$V        \\ 
(6)   & (5)   & conv1d(C$\rightarrow$C, $1\times1$), Relu & N$\times$C$\times$V        \\ 
(7)   & (1), (2)             & SPAdaIN3 (C=C)        & N$\times$C$\times$V        \\ 
(8)   & (7)             & conv1d(C$\rightarrow$C, $1\times1$), Relu        & N$\times$C$\times$V        \\ 
(9)   & (5), (8)              & Add            & N$\times$C$\times$V        \\ \hline
\end{tabular}
\end{adjustbox}
\caption{The network architecture for SPAdaIN Res-Block.}
\label{tab::spadain_resblock}
\end{table}

\begin{table}[h]
\begin{tabular}{|c|c|c|c|}
\hline
Index & Inputs     & Operation  & Output shape \\ \hline
(1)   & Input      & Identity Mesh     & N$\times$3$\times$V        \\ 
(2)   & Input & Input Features    & N$\times$C$\times$V        \\ 
(3)   & (1)  & conv1d(3$\rightarrow$C, $1\times1$) & N$\times$C$\times$V        \\ 
(4)   & (1)  & conv1d(3$\rightarrow$C, $1\times1$) & N$\times$C$\times$V        \\ 
(5)   & (2)  & Instance Norm     & N$\times$C$\times$V        \\ 
(6)   & (3), (5)          & Multiply    & N$\times$C$\times$V        \\ 
(7)   & (4), (6)          & Add & N$\times$C$\times$V        \\ \hline
\end{tabular}
\caption{The network architecture for SPAdaIN unit.}
\label{tab::spadain}
\end{table}

\section{Data Preparation}
We prepare our training and testing data using SMPL~\cite{loper2015smpl} model.
SMPL~\cite{loper2015smpl} model has 10 morphology parameters controlling the shape and 24 sets of joint parameters controlling the pose. For shape parameters, we randomly sample from the parameter space. For pose parameters, each set of parameters has three sub-parameters represented as a tuple $(x, y, z)$, indicating rotated joint angle around x-axis, y-axis and z-axis respectively. In order to generate natural looking poses, we constrain the rotation angle of the joints according to what human joints can physically reach. Then we sample from the constrained angle space. The details of the range can be seen in Tab.~\ref{tab::data}.
\section{Comparison to Baselines}
In this section, we design and evaluate some competitive baselines.
\subsection{Comparison to Skeleton Pose Driven Approach}
We compare our method with skeleton-based skinning shape deformation. 
We first extract human pose skeleton from both the pose and identity meshes by fitting an SMPL~\cite{loper2015smpl} model. 
We take the T-pose SMPL as the initialization, and update the SMPL parameters through gradient descent using LBFGS~\cite{liu1989limited}.
We use the joints of this fitted model as the key points of our skeleton representation.  
We then calculate the binding weights of LBS (Linear Blend Skinning)~\cite{elrond79Pinocchio,lewis2000pose,jacobson2012fast,jacobson2011bounded} using tools from Baran \etal \cite{baran2007automatic}. 
After that, we transform the identity skeleton to the pose skeleton. Since the skeleton joints of SMPL model assemble a kinematic tree, we calculate the transformation matrix between two skeletons according to the connection relationship of the joints through the local coordinate system.
Finally, we recover the mesh from skeleton using the binding weights computed before.

\begin{figure*}[h!]
\begin{center}
\includegraphics[width=1\linewidth]{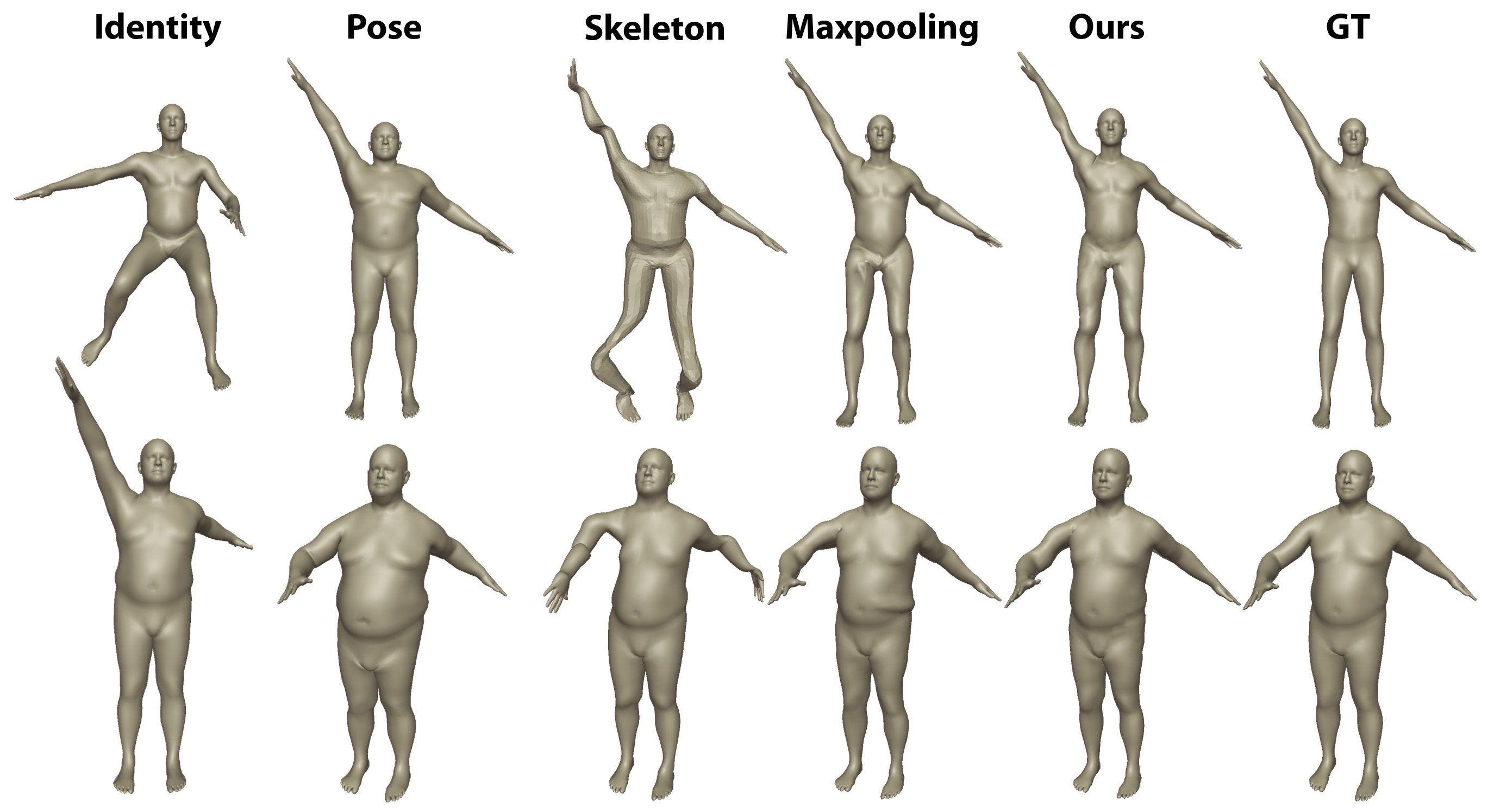}
\end{center}
   \caption{\textbf{Qualitative comparison to other baselines.} From left to right, we show in each row: input identity mesh, input pose mesh, the results of skeleton pose driven approach, the results of max pooling method, our results and the ground truth. We have more accurate results.}
\label{fig:compare}
\end{figure*}

We show the quantitative result in Tab.~\ref{tab:baseline}. According to the table, the skeleton based approach cannot perform as well as our method due to the accumulated error at each stage. 
Particularly, this method has trouble dealing with varying limb length caused by body shape variations.
Qualitative evaluation is shown in Fig.~\ref{fig:compare}. The skeleton based deformation approach often produces artifacts near joint points, due to different limb lengths.

\begin{table*}[tbh]
\centering
\begin{tabular}{|c|c|c|c|}
\hline
\textbf{Index} & \textbf{Inputs} & \textbf{Operation} & \textbf{Output Shape} \\ \hline
(1)   & Input   & Identity Mesh & N$\times$3$\times$V    \\ 
(2)   & Input   & Pose Mesh     & N$\times$3$\times$V    \\ 
(3)   & (1)     & conv1d(3$\rightarrow$64, $1\times 1$) & N$\times$64$\times$V   \\ 
(4)   & (3)     & Instance Norm, Relu & N$\times$64$\times$V   \\ 
(5)   & (4)     & conv1d(64$\rightarrow$128, $1\times 1$) & N$\times$128$\times$V  \\ 
(6)   & (5)     & Instance Norm, Relu & N$\times$128$\times$V  \\ 
(7)   & (6)     & conv1d(128$\rightarrow$1024, $1\times 1$) & N$\times$1024$\times$V \\ 
(8)   & (7)     & Instance Norm, Relu & N$\times$1024$\times$V \\ 
(9)   & (2), (8) & Concatenate            & N$\times$1027$\times$V \\ 
(10)  & (9)     & conv1d(1027$\rightarrow$1027, $1\times 1$) & N$\times$1027$\times$V \\ 
(11)  & (10)    & SPAdaIN ResBlk 1 (C=1027)       & N$\times$1027$\times$V \\ 
(12)  & (11)    & conv1d(1027$\rightarrow$513, $1\times 1$)  & N$\times$513$\times$V  \\ 
(13)  & (12)    & SPAdaIn ResBlk 2 (C=513)       & N$\times$513$\times$V  \\ 
(14)  & (13)    & conv1d(513$\rightarrow$256, $1\times 1$)   & N$\times$256$\times$V  \\ 
(15)  & (14)    & SPAdaIN ResBlk 3 (C=256)       & N$\times$256$\times$V  \\ 
(16)  & (15)    & conv1d(256$\rightarrow$3, $1\times 1$)     & N$\times$3$\times$V    \\ 
(17)  & (16)    & tanh      & N$\times$3$\times$V    \\ 
\hline
\end{tabular}
\caption{The network architecture for our full model.}
\label{tab::full_network}
\end{table*} 

\subsection{Comparison to Compact Pose Feature}
We also create a strong deep learning baseline.
Instead of maintaining the per-vertex feature on the pose mesh, we apply a global max pooling as suggested in PointNet to extract a compact global pose feature.
This feature is then concatenated with each vertex in the identity mesh, and further fed into the decoder.
Note that we need to remove the first instance normalization in the decoder to make this work, otherwise the instance normalization would whitening all the pose feature as they are exactly the same on all the vertices.

The quantitative result is shown in Tab.~\ref{tab:baseline}. As can be seen, this baseline works much better than the skeleton based deformation, but not as good as our method.
One possible reason could be that the global max pooling may drop some fine-grained information from the pose mesh which is helpful for pose transfer.

\begin{table}[h]
\centering
\begin{tabu} to \columnwidth {X[1.5c]X[c]X[c]X[c]}
\toprule
\multirow{3}{*}{Parameter Index} &
\multicolumn{3}{c}{Rotation Degree of Axes}\\\tabuphantomline
\cmidrule(lr){2-4}
   & x-axis   & y-axis   & z-axis   \\ \midrule \midrule 
1  & (-2,2)   & (-2,2)   & (-2,2)   \\ 
2  & (-90,0)  & 0        & (0,40)   \\ 
3  & (-90,0)  & 0        & (-40,0)  \\ 
4  & (-1,1)   & (-1,1)   & (-1,1)   \\ 
5  & (0,100)  & 0        & 0        \\ 
6  & (0,100)  & 0        & 0        \\ 
7  & (-1,1)   & (-1,1)   & (-1,1)   \\ 
8  & (-10,10) & (-10,10) & (-1,1)   \\ 
9  & (-10,10) & (-10,10) & (-1,1)   \\ 
10 & (-1,1)   & (-1,1)   & (-1,1)   \\ 
11 & (-1,1)   & (-1,1)   & (-1,1)   \\ 
12 & (-1,1)   & (-1,1)   & (-1,1)   \\ 
13 & (-3,3)   & (-3,3)   & (-3,3)   \\ 
14 & 0        & (-30,30) & (-30,30) \\ 
15 & 0        & (-30,30) & (-30,30) \\ 
16 & (-3,3)   & (-3,3)   & (-3,3)   \\ 
17 & 0        & (-30,30) & (-30,30) \\ 
18 & 0        & (-30,30) & (-30,30) \\ 
19 & 0        & (-60,0)  & 0        \\ 
20 & 0        & (0,60)   & 0        \\ 
21 & (-10,10) & (-10,10) & (-10,10) \\ 
22 & (-10,10) & (-10,10) & (-10,10) \\ 
23 & (-5,5)   & (0,10)   & (-10,0)  \\ 
24 & (-5,5)   & (-10,0)  & (0,10)   \\ 
\bottomrule
\end{tabu}
\caption{\textbf{Pose parameters preparation.} Human posture can be easily adjusted by rotating 24 key joints represented as parameter index. We give more details of the range of angles of each pose parameter. We randomly sample in this pose space to generate our input data.}
\label{tab::data}
\end{table}

\begin{table}[h]
\centering
\begin{tabu} to \columnwidth {X[2c]X[c]X[c]X[c]}
\toprule
\multirow{3}{*}{Pose Source} &
\multicolumn{3}{c}{PMD~$\downarrow$~($\times 10^{-4}$)}\\
\tabuphantomline
\cmidrule(lr){2-4}
& skeleton & maxpooling & ours \\
\midrule 
\midrule 
seen-pose & 27.4 & 2.1 & 1.1 \\
unseen-pose & 31.1 & 12.7 & 9.3 \\
\bottomrule
\end{tabu}
\caption{\textbf{Quantitative comparison to other baselines.}}
\label{tab:baseline}
\end{table}

\begin{figure}[h]
	\centering
	\includegraphics[width=\columnwidth]{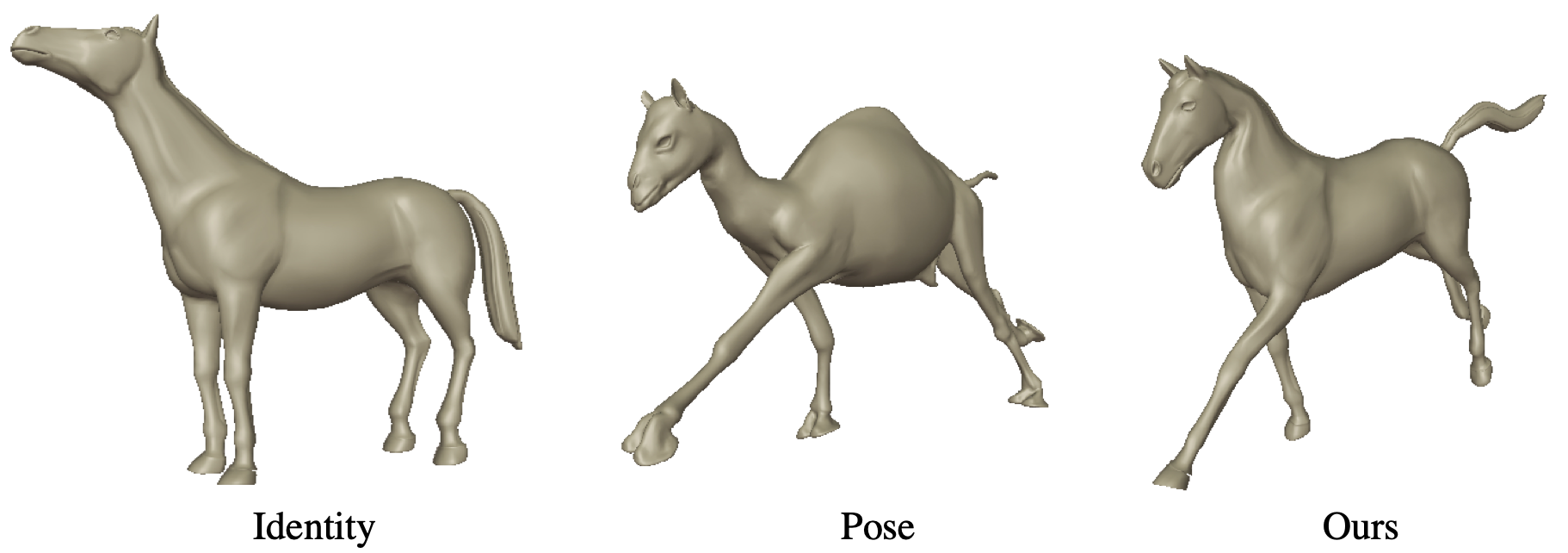}
	\caption{\textbf{Camel and horse pose transfer.}}
	\label{fig:non-human}
\end{figure}


\section{More Qualitative Results}
In this section, we show more qualitative results to demonstrate the robustness and generalization capability of our system. 

\subsection{Invariance to Vertex Order}
\begin{figure*}[b]
\begin{center}
\includegraphics[width=1\linewidth]{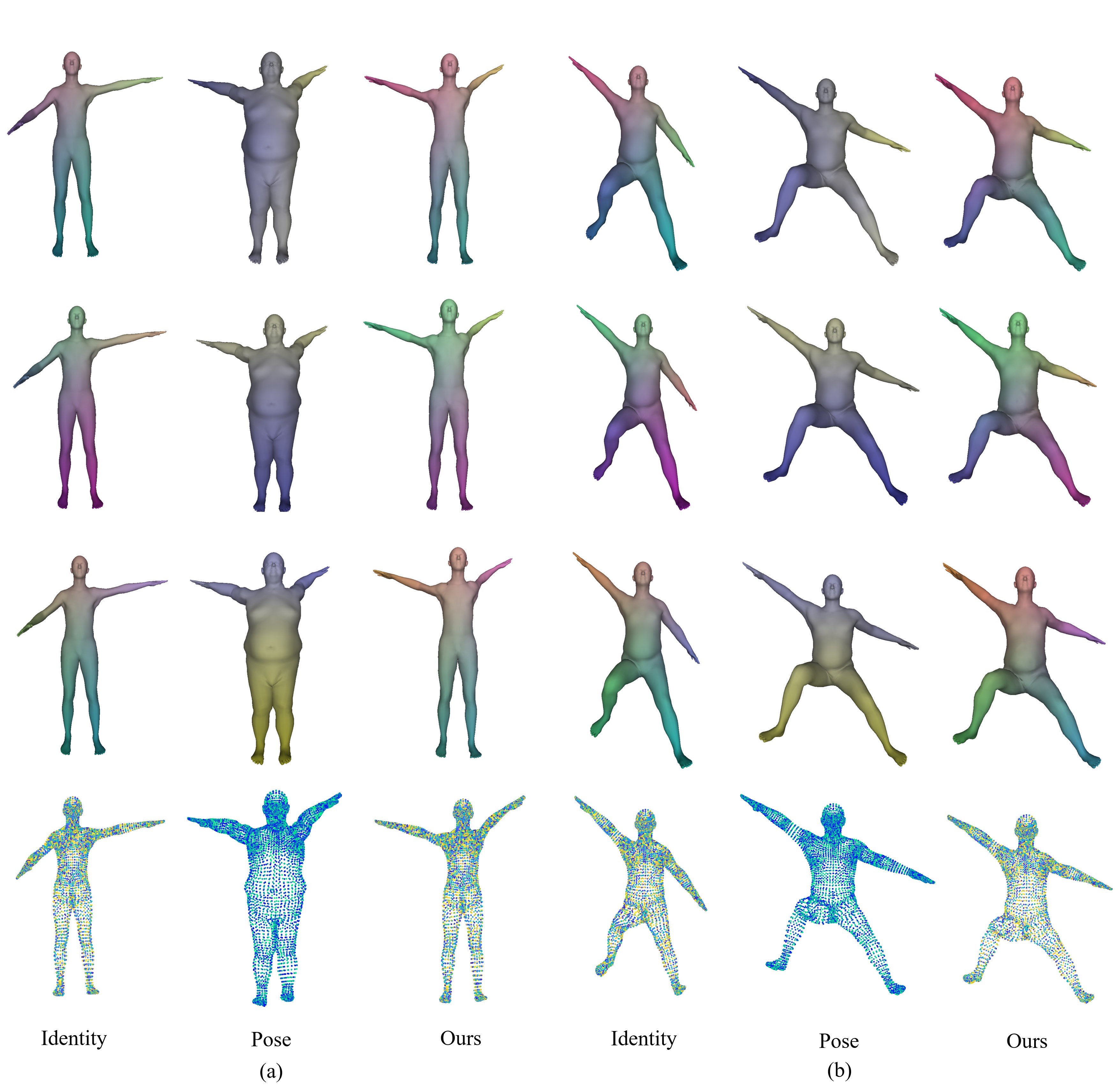}
\end{center}
   \caption{\textbf{More qualitative results of different vertex order.} Each row represents one vertices order with respect to input identity mesh, pose mesh and our results. Our results are consistent with the identity mesh. Different orders of the inputs do not affect the output visually. For all meshes, the vertex color represents the vertex order. For the first 3 rows, vertices are colored according to the \textit{index$\rightarrow$color mapping}, and for the last row, vertices are colored according to the \textit{value of the index}.}
\label{fig:dif_order2}
\end{figure*}

To the best of our knowledge, our model is the first one that achieves permutation invariance on the order of vertices in both input meshes. That says, the identity mesh can be provided in arbitrary pose and vertex order.
We verify the model behavior with random permutation, and the results are shown in Fig.~\ref{fig:dif_order2}.
For each example one the left and right, we randomly shuffle the vertex order in both the identity and pose mesh, and feed them into the same network (we use the color to encode the vertex order).
As can be seen, our network successfully produces visually the same target mesh with correct identity and pose.
Note that for each random shuffle, the output vertex order is the same as the identity mesh.
This indicates that the deformed mesh are point-wise aligned with the initial identity mesh, which can be very useful for many graphics applications, \eg texture transfer.

\subsection{Robustness to Pose Mesh Noise}
\begin{figure*}[t]
\begin{center}
\includegraphics[width=0.9\linewidth]{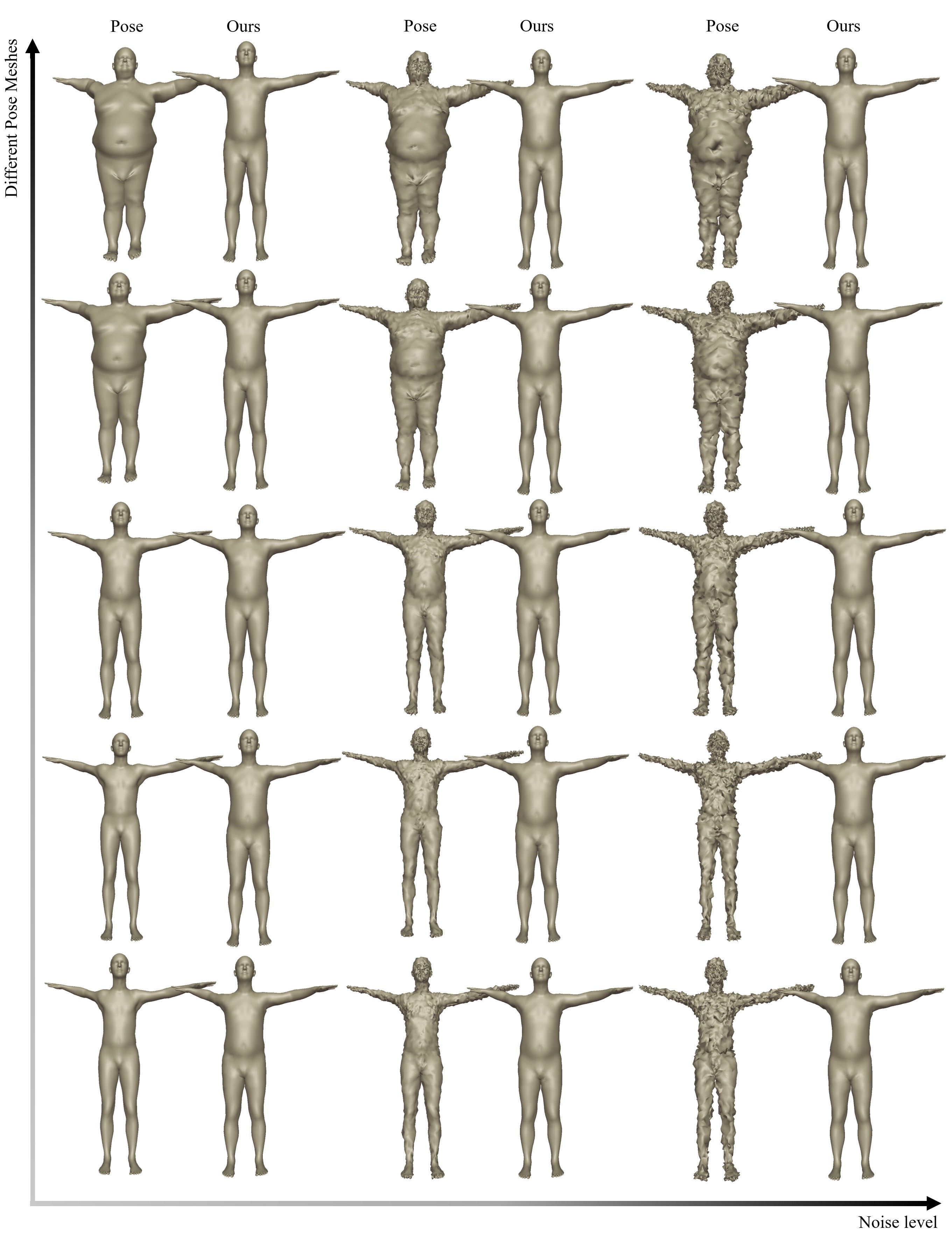}
\end{center}
   \caption{\textbf{Robustness to different pose meshes.} The pose meshes in the same pose with different shape or the pose meshes with different level of noise would not influence the result. Our method can produce similar and correct output.}
\label{fig:robustness}
\end{figure*}

We test the robustness of our system given noisy mesh.
In Fig.~\ref{fig:robustness}, we provide our model pose meshes in the same pose but with different shape and noise level.
Our network successfully extracts the correct pose information and produces final output mesh in the correct pose.

\subsection{Generalization to New Identity}
\begin{figure*}[t]
\begin{center}
\includegraphics[width=1\linewidth]{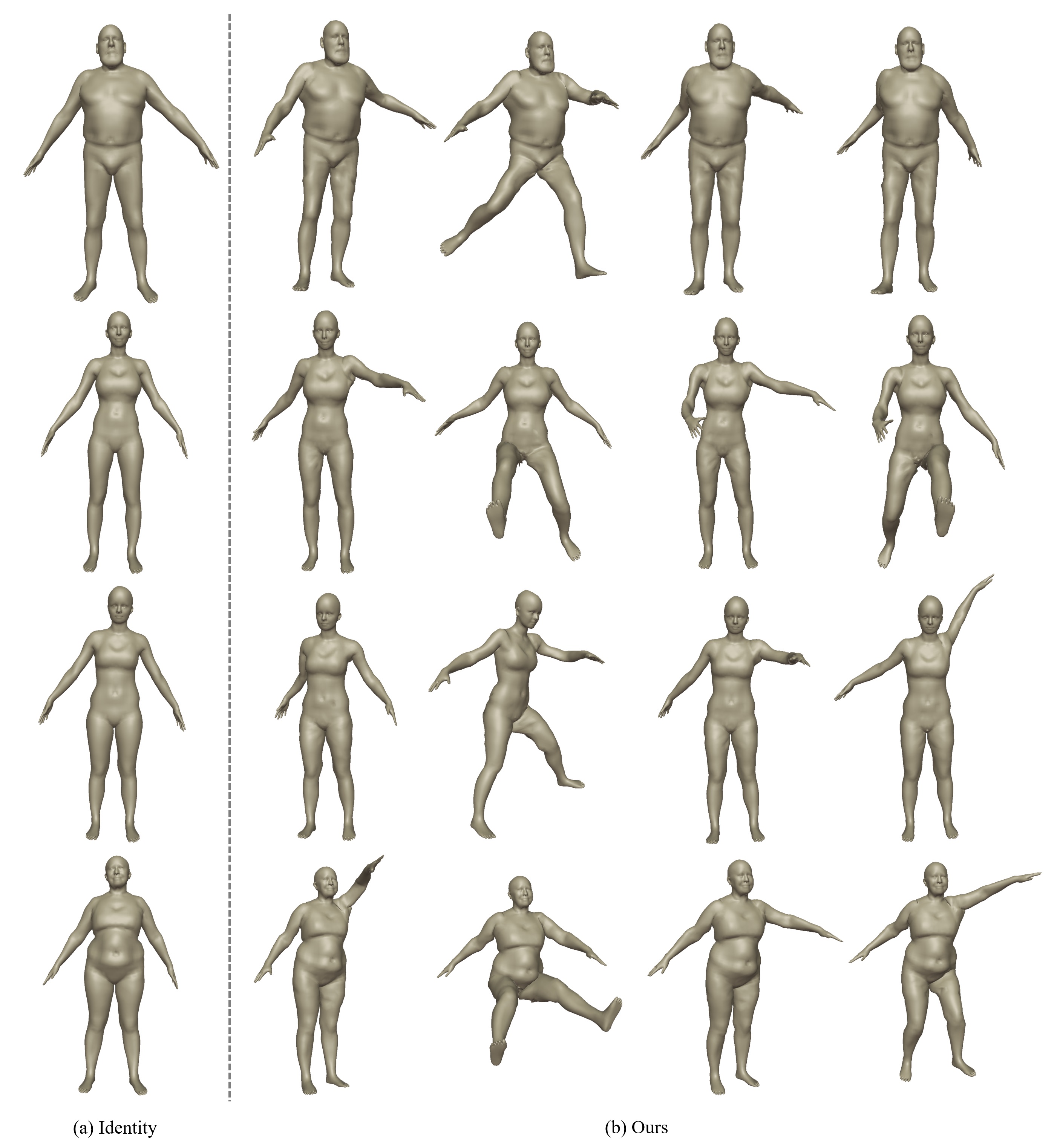}
\end{center}
   \caption{\textbf{More examples of identity mesh from FAUST~\cite{Bogo_2014_CVPR}.} (a) Identity input meshes. (b) Output meshes using our methods. Our method can deform the identity mesh to various poses with good visual quality.}
\label{fig:faust}
\end{figure*}

We also test the generalization capability of our model to unseen identities, especially those non-SMPL model meshes.
In Fig.~\ref{fig:faust}, we show more results on identity meshes from FAUST dataset~\cite{Bogo_2014_CVPR}. 
Our model generalizes to these meshes automatically without any finetune.
Features that not measured by SMPL, such as the mustache of the man in the first row, are successfully maintained.

\begin{figure*}[t]
\begin{center}
\includegraphics[width=1\linewidth]{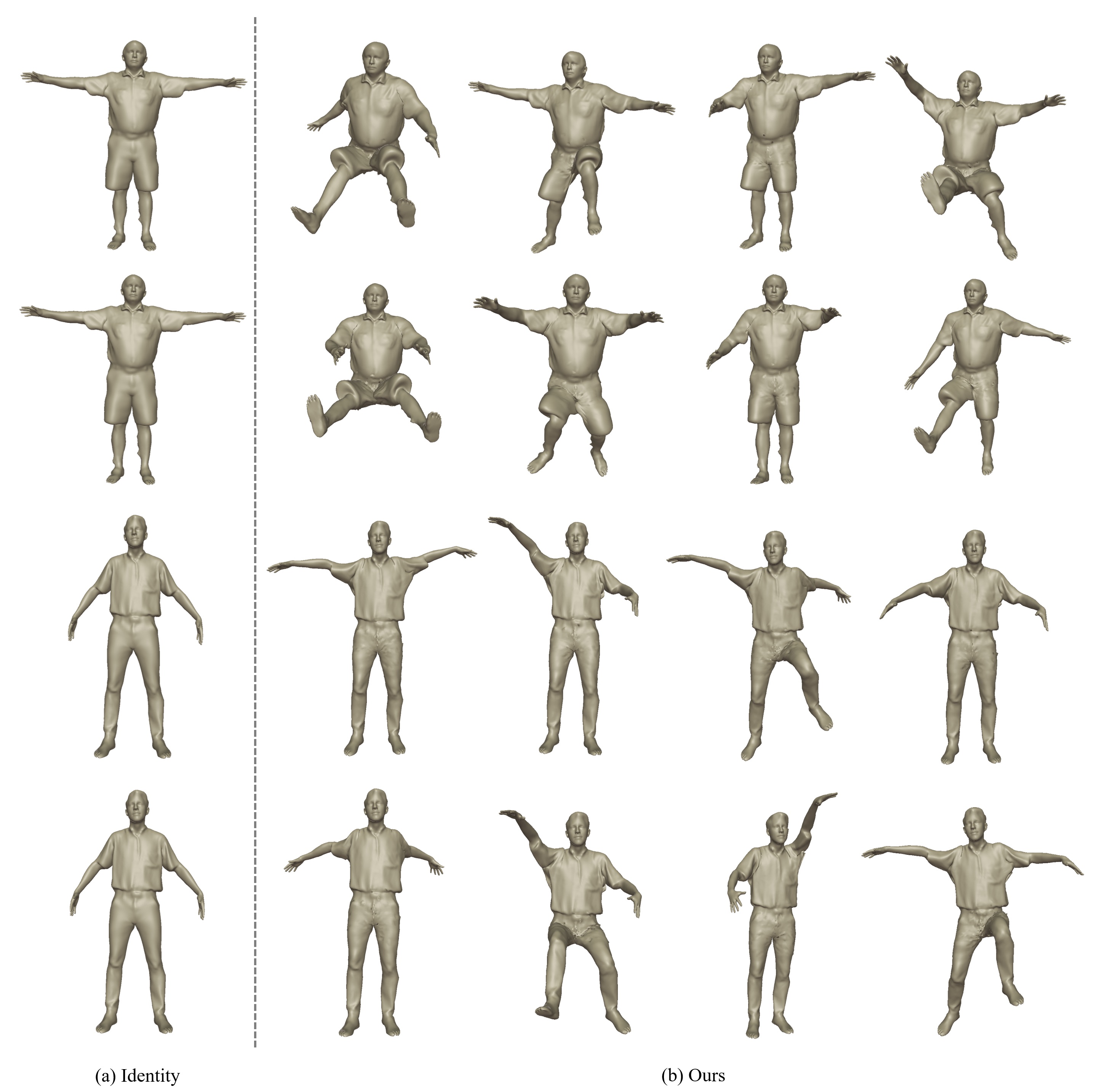}
\end{center}
   \caption{\textbf{More examples of identity mesh from MG-dataset~\cite{bhatnagar2019mgn}.} (a) Identity input meshes (b) Output meshes using our methods. Though the appearance of MG-dataset~\cite{bhatnagar2019mgn} wearing clothes is quite different from meshes of SMPL, our method can still produce very good results.}
\label{fig:cloth}
\end{figure*}

We also try more challenging cases using meshes from MG-dataset~\cite{bhatnagar2019mgn}.
The meshes in this dataset contains apparel, which are more different with SMPL meshes compared to those from the FAUST~\cite{Bogo_2014_CVPR}.
As can be seen in Fig.~\ref{fig:cloth}, though with some small artifacts, our model still maintains the identity, \ie person and apparel, correctly.

\subsection{Our Results on Seen and Unseen Poses}
\begin{figure*}[t]
\begin{center}
\includegraphics[width=0.8\linewidth]{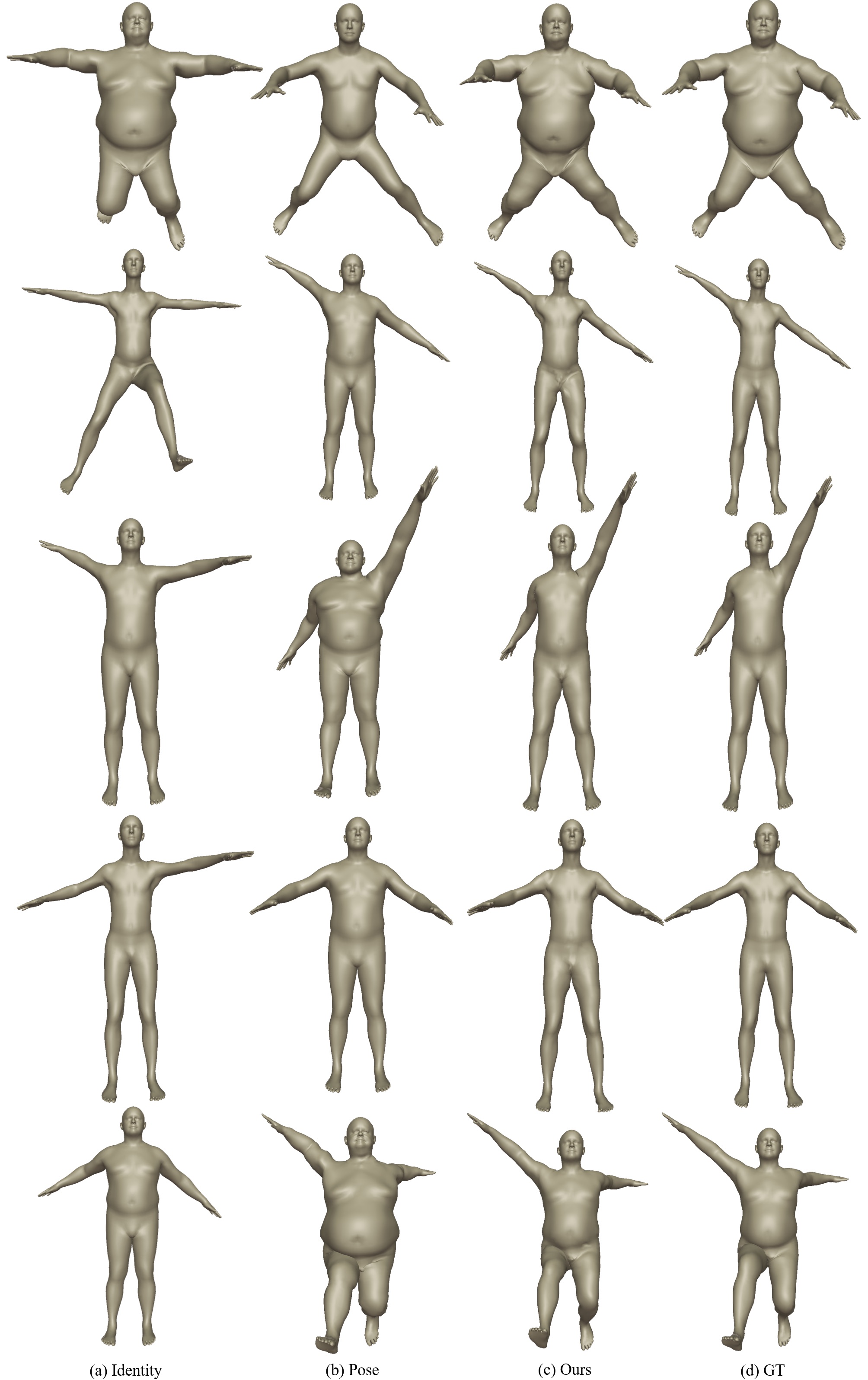}
\end{center}
   \caption{\textbf{More examples of seen poses.} From left to right, we show in each row: input identity mesh, input pose mesh, the results of ours and the ground truth. }
\label{fig:seen}
\end{figure*}

\begin{figure*}[t]
\begin{center}
\includegraphics[width=0.8\linewidth]{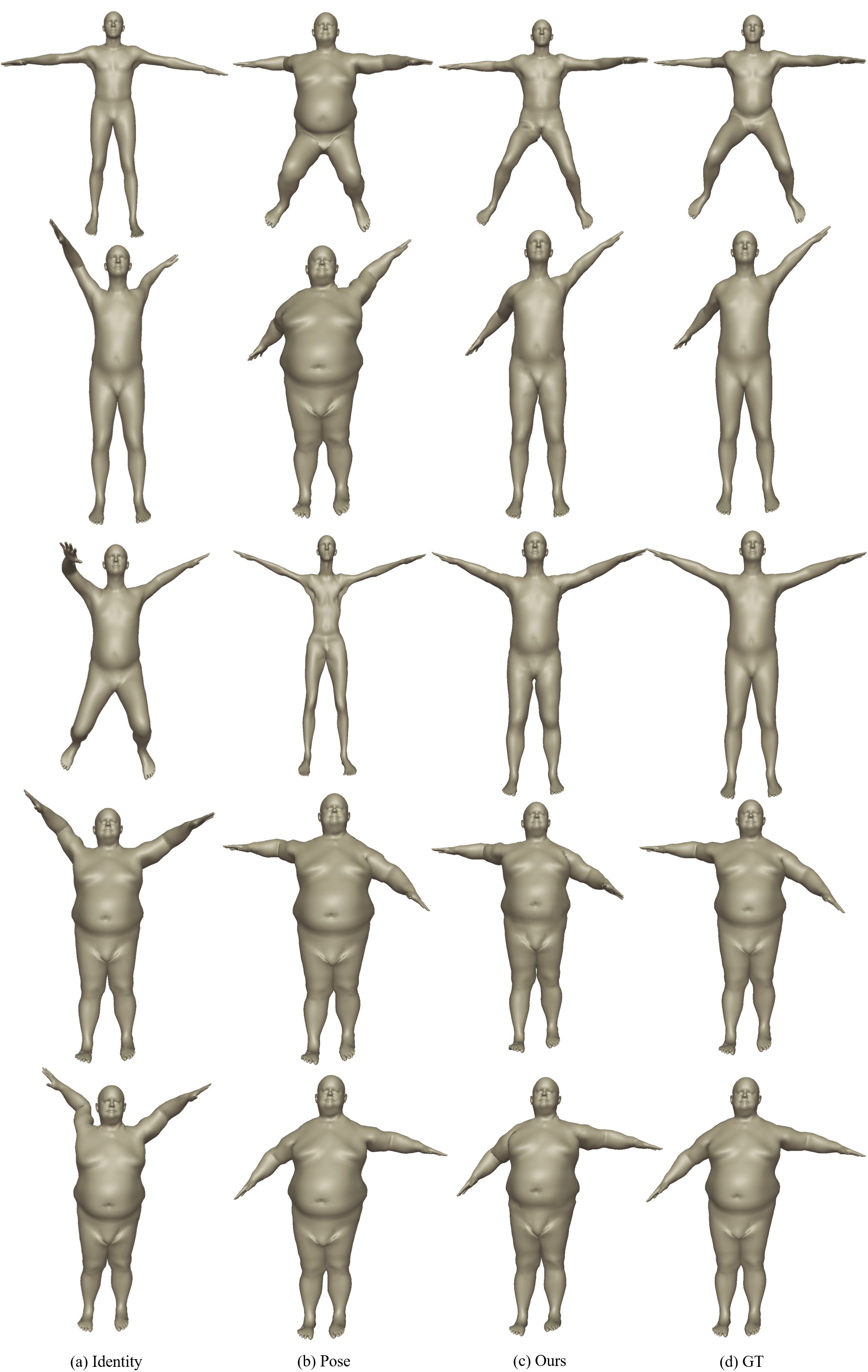}
\end{center}
   \caption{\textbf{More examples of unseen poses.} From left to right, we show in each row: input identity mesh, input pose mesh, the results of ours and the ground truth. }
\label{fig:unseen}
\end{figure*}

We show more qualitative results of our model on seen and unseen poses in Fig.~\ref{fig:seen} and Fig.~\ref{fig:unseen} respectively.

\subsection{Our Results on Non-Human Models}
In the end, we show the results of our model on transferring pose from camel to horse in Fig.~\ref{fig:non-human}, by training on the animal dataset~\cite{sumner2004deformation}. 
We adopt the compact pose feature encoder to handle different vertices number between identity mesh and pose mesh, and then using our decoder to transfer pose for non-human meshes.
Even though we specifically focus on human, our model also works for non-human meshes but require domain-specific training.

{\small
\bibliographystyle{ieee_fullname}
\bibliography{egbib.bib}
}

\end{document}

%% file: math_commands.tex
\usepackage{amsmath,amsfonts,bm}

\def\eqref#1{equation~\ref{#1}}

\def\1{\bm{1}}

\def\mF{{\bm{F}}}

\def\mM{{\bm{M}}}

\def\mZ{{\bm{Z}}}

\DeclareMathAlphabet{\mathsfit}{\encodingdefault}{\sfdefault}{m}{sl}
\SetMathAlphabet{\mathsfit}{bold}{\encodingdefault}{\sfdefault}{bx}{n}

